\documentclass[11pt]{article}

\usepackage[final]{acl}

\usepackage{times}
\usepackage{latexsym}

\usepackage[T1]{fontenc}
\usepackage[utf8]{inputenc}

\usepackage{microtype}

\usepackage{inconsolata}

\usepackage{graphicx}

\usepackage{amsmath}
\usepackage{amssymb}

\usepackage{algorithm}
\usepackage{algorithmic}

\usepackage{tcolorbox}
\usepackage{xcolor}
\usepackage{colortbl}
\usepackage{booktabs}
\usepackage{multirow}
\usepackage{enumitem}

\usepackage{tikz}
\usetikzlibrary{shapes,shapes.misc,arrows,positioning,fit,backgrounds,calc,decorations.pathreplacing}

\usepackage{pifont}

\usepackage{fontawesome5}

\definecolor{highlightblue}{RGB}{230,242,255}
\definecolor{contributioncolor}{RGB}{0,100,180}

\definecolor{abstractcolor}{RGB}{41,128,185}    % Blue for abstraction (phi)
\definecolor{progresscolor}{RGB}{39,174,96}     % Green for progression (psi)
\definecolor{baselinecolor}{RGB}{192,57,43}     % Red for baseline issues

\newcommand{\method}{\textsc{Stratagem}}

\title{\textsc{Stratagem}: Learning Transferable Reasoning via \\ Trajectory-Modulated Game Self-Play}

\author{
Xiachong Feng\textsuperscript{1}\thanks{~Equal contribution.},
Deyi Yin\textsuperscript{2}\footnotemark[1],
Xiaocheng Feng\textsuperscript{2}\thanks{~Corresponding author.},
Yi Jiang\textsuperscript{2},
Libo Qin\textsuperscript{3},
Yangfan Ye\textsuperscript{2},
Lei Huang\textsuperscript{2},\\
\textbf{Weitao Ma}\textsuperscript{2},
\textbf{Qiming Li}\textsuperscript{2},
\textbf{Yuxuan Gu}\textsuperscript{2},
\textbf{Bing Qin}\textsuperscript{2},
\textbf{Lingpeng Kong}\textsuperscript{1}\footnotemark[2]\\
\textsuperscript{1}The University of Hong Kong \quad
\textsuperscript{2}Harbin Institute of Technology \quad
\textsuperscript{3}Harbin Institute of Technology, Shenzhen\\
\texttt{fengxc@hku.hk, xcfeng@ir.hit.edu.cn, lpk@cs.hku.hk}
}

\begin{document}
\maketitle
\begin{abstract}
Games offer a compelling paradigm for developing general reasoning capabilities in language models, as they naturally demand strategic planning, probabilistic inference, and adaptive decision-making. However, existing self-play approaches rely solely on terminal game outcomes, providing no mechanism to distinguish transferable reasoning patterns from game-specific heuristics. We present \method{}, which addresses two fundamental barriers to reasoning transfer: \textit{domain specificity}, where learned patterns remain anchored in game semantics, and \textit{contextual stasis}, where static game contexts fail to cultivate progressive reasoning. \method{} selectively reinforces trajectories exhibiting abstract, domain-agnostic reasoning through a \textit{Reasoning Transferability Coefficient}, while incentivizing adaptive reasoning development via a \textit{Reasoning Evolution Reward}. Experiments across mathematical reasoning, general reasoning, and code generation benchmarks demonstrate substantial improvements, with particularly strong gains on competition-level mathematics where multi-step reasoning is critical. Ablation studies and human evaluation confirm that both components contribute to transferable reasoning.\footnote{Code: \href{https://github.com/ydyyyy/Stratagem}{\faGithub\ Stratagem}.}
\end{abstract}

\section{Introduction}

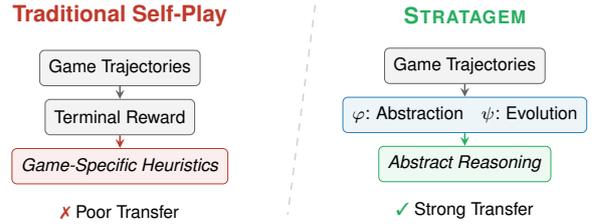
\begin{figure}[t]
\centering
\resizebox{0.48\textwidth}{!}{%
\begin{tikzpicture}[
    node distance=0.25cm,
    box/.style={rectangle, rounded corners=2pt, draw=#1, fill=#1!8, minimum width=2cm, minimum height=0.5cm, align=center, font=\scriptsize\sffamily},
    arrow/.style={->, >=stealth, semithick, #1},
]

% Colors
\definecolor{badcolor}{RGB}{192,57,43}
\definecolor{goodcolor}{RGB}{39,174,96}
\definecolor{neutralcolor}{RGB}{100,100,100}
\definecolor{phicolor}{RGB}{41,128,185}

% === Left Panel ===
\node[font=\small\sffamily\bfseries, badcolor] (left-title) {Traditional Self-Play};
\node[box=neutralcolor, below=0.2cm of left-title] (game1) {Game Trajectories};
\node[box=neutralcolor, below=0.18cm of game1] (reward1) {Terminal Reward};
\node[box=badcolor, below=0.18cm of reward1] (pattern1) {\textit{Game-Specific Heuristics}};
\node[below=0.15cm of pattern1, font=\scriptsize\sffamily] (transfer1) {\textcolor{badcolor}{\ding{55}} Poor Transfer};

\draw[arrow=neutralcolor] (game1) -- (reward1);
\draw[arrow=badcolor] (reward1) -- (pattern1);

% === Right Panel ===
\node[font=\small\sffamily\bfseries, goodcolor, right=2.2cm of left-title] (right-title) {\textsc{Stratagem}};
\node[box=neutralcolor, below=0.2cm of right-title] (game2) {Game Trajectories};
\node[box=phicolor, below=0.18cm of game2] (modulate) {$\varphi$: Abstraction\quad $\psi$: Evolution};
\node[box=goodcolor, below=0.18cm of modulate] (pattern2) {\textit{Abstract Reasoning}};
\node[below=0.15cm of pattern2, font=\scriptsize\sffamily] (transfer2) {\textcolor{goodcolor}{\ding{51}} Strong Transfer};

\draw[arrow=neutralcolor] (game2) -- (modulate);
\draw[arrow=goodcolor] (modulate) -- (pattern2);

% Separator
\draw[dashed, neutralcolor!30, line width=0.6pt] ($(left-title.east)!0.5!(right-title.west)+(0,0.15)$) -- ($(transfer1.east)!0.5!(transfer2.west)+(0,-0.05)$);

\end{tikzpicture}%
}
\caption{Traditional self-play learns game-specific heuristics from terminal rewards. \method{} modulates trajectory advantages via abstraction ($\varphi$) and evolution ($\psi$), selectively reinforcing transferable reasoning.}
\label{fig:teaser}
\end{figure}

% Paragraph 1: AI in games - opportunity for AGI
Games have long served as a proving ground for artificial intelligence, offering structured environments where complex reasoning emerges from simple rules \citep{silver2016mastering, berner2019dota, vinyals2019grandmaster}. Beyond serving as evaluation benchmarks, games provide a unique opportunity for cultivating general reasoning capabilities: they demand strategic planning, probabilistic inference, and adaptive decision-making, all cognitive skills that underpin intelligent behavior across diverse domains \citep{xu2024survey, hu2025lmgame}. This observation has motivated a growing body of work exploring games as training environments for language models \citep{Hu2024ASO, Tong2025GameRLSM, Xie2025PlayTG}, premised on the hypothesis that reasoning patterns developed through gameplay may transfer to downstream tasks such as mathematical problem-solving and code generation.

% Paragraph 2: Self-play importance, history, SPIRAL and its drawbacks
Self-play has emerged as a promising paradigm within this agenda, enabling models to improve through competitive interaction without requiring curated datasets \citep{zhang2024survey, zhao2025absolute}. Historical successes in game-playing AI, from AlphaGo \citep{silver2016mastering} to OpenAI Five \citep{berner2019dota}, demonstrate that self-play can produce superhuman performance in specific domains. Recent work has extended this paradigm to language models: SPIRAL \citep{liu2025spiral} trains LLMs through self-play on text-based zero-sum games, showing that game-derived rewards can improve reasoning capabilities. However, SPIRAL relies on terminal game outcomes (win/loss) to provide learning signals, offering no explicit mechanism to identify or reinforce reasoning patterns that transfer beyond game-specific contexts. As a result, models may learn to win games through domain-specific heuristics (e.g., ``King beats Queen'') that fail to generalize, while transferable reasoning (e.g., ``enumerate cases and compute expected value'') receives no preferential reinforcement.

% Paragraph 3: Our method - STRATAGEM
To address this limitation, we propose \method{} (\textbf{S}elf-Play \textbf{TR}ajectory \textbf{A}dvan\textbf{T}age \textbf{A}ctivated \textbf{G}am\textbf{E} Lear\textbf{M}ing), which learns transferable reasoning by selectively reinforcing trajectories that exhibit domain-agnostic and adaptive reasoning patterns. Our key insight is that transfer requires addressing two fundamental challenges: \textit{domain specificity}, where game-learned patterns remain anchored in game semantics rather than abstract principles; and \textit{contextual stasis}, where static game environments fail to cultivate the progressive reasoning needed for evolving problem contexts. \method{} tackles both challenges by modulating trajectory advantages with two complementary signals: a \textit{Reasoning Transferability Coefficient} ($\varphi$) that measures the abstraction level of reasoning patterns, and a \textit{Reasoning Evolution Reward} ($\psi$) that incentivizes reasoning that deepens and adapts across turns. By multiplicatively scaling advantage based on transferability and additively rewarding reasoning evolution, \method{} ensures that only trajectories demonstrating both abstract reasoning and progressive development receive maximal reinforcement.

% Paragraph 4: Experiments
We evaluate \method{} on benchmarks spanning mathematical reasoning, general reasoning, and code generation. Training on three text-based games using Qwen3-4B-Base, \method{} achieves consistent improvements across all categories, with strong gains on competition-level mathematics where multi-step reasoning is critical. Ablation studies confirm that both modulation components contribute meaningfully, while human evaluation validates that \method{} produces more abstract and progressive reasoning.

% Paragraph 5: Contributions
Our contributions are:
\begin{itemize}[leftmargin=*, topsep=2pt, itemsep=1pt]
\item We identify \textit{domain specificity} and \textit{contextual stasis} as two fundamental barriers to reasoning transfer in game-based self-play, and propose \method{} to address both through selective trajectory advantage modulation.

\item We introduce the Reasoning Transferability Coefficient ($\varphi$) that quantifies abstraction level, and Reasoning Evolution Reward ($\psi$) that incentivizes progressive reasoning development.

\item We demonstrate strong transfer across mathematical reasoning, general reasoning, and code generation, with notable gains on competition-level problems requiring multi-step reasoning.
\end{itemize}
\section{Preliminaries}

\subsection{Task Formulation}

We formulate multi-turn reasoning as a turn-level Markov Decision Process (MDP) $\mathcal{M} = (\mathcal{S}, \mathcal{A}, T, r, \gamma)$, where states $s \in \mathcal{S}$ represent complete contexts (e.g., game configurations) and actions $a \in \mathcal{A}$ correspond to full responses rather than individual tokens (see Appendix~\ref{app:task_formulation} for extended background). At each turn $t$, the model generates response $y_t$ containing reasoning $c_t$ and executable action $a_t$.

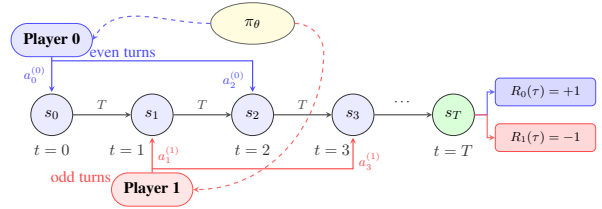
\begin{figure}[t]
\centering
\resizebox{\columnwidth}{!}{%
\begin{tikzpicture}[
    % Node styles
    state/.style={circle, draw=black!80, fill=blue!8, minimum size=10mm, font=\normalsize},
    terminal/.style={circle, draw=black!80, fill=green!15, minimum size=10mm, font=\normalsize},
    player/.style={rounded rectangle, minimum width=22mm, minimum height=8mm, font=\normalsize\bfseries},
    player0/.style={player, draw=blue!70, fill=blue!10},
    player1/.style={player, draw=red!70, fill=red!10},
    policy/.style={ellipse, draw=black!60, fill=yellow!15, minimum width=20mm, minimum height=10mm, font=\normalsize},
    reward/.style={rectangle, rounded corners=3pt, font=\small, inner sep=5pt, minimum width=24mm},
    % Arrow styles
    action/.style={->, >=stealth, thick},
    trans/.style={->, >=stealth, thick, black!70},
    policy_arrow/.style={->, >=stealth, dashed, thick},
]

% States
\node[state] (s0) at (0, 0) {$s_0$};
\node[state] (s1) at (2.4, 0) {$s_1$};
\node[state] (s2) at (4.8, 0) {$s_2$};
\node[state] (s3) at (7.2, 0) {$s_3$};
\node[terminal] (sT) at (9.6, 0) {$s_T$};

% Players
\node[player0] (p0) at (0, 1.8) {Player 0};
\node[player1] (p1) at (2.4, -1.8) {Player 1};

% Shared policy
\node[policy] (pi) at (4.8, 2.2) {$\pi_\theta$};

% Policy connections - route red line around to avoid overlap
\draw[policy_arrow, blue!60] (pi.west) to[out=180, in=60] (p0.east);
\draw[policy_arrow, red!60] (pi.east) to[out=0, in=90] (6.5, 1.5) to[out=-90, in=0] (p1.east);

% Turn 0: Player 0 acts
\draw[action, blue!70] (p0.south) -- node[left, font=\small] {$a_0^{(0)}$} (0, 0.55);
\draw[trans] (s0) -- node[above, font=\small] {$T$} (s1);

% Turn 1: Player 1 acts
\draw[action, red!70] (p1.north) -- node[right, font=\small] {$a_1^{(1)}$} (2.4, -0.55);
\draw[trans] (s1) -- node[above, font=\small] {$T$} (s2);

% Turn 2: Player 0 acts - label moved down
\draw[action, blue!70] (0, 1.3) -| (4.8, 0.55);
\node[font=\small, blue!70] at (4.35, 0.85) {$a_2^{(0)}$};
\draw[trans] (s2) -- node[above, font=\small] {$T$} (s3);

% Turn 3: Player 1 acts
\draw[action, red!70] (2.4, -1.3) -| node[below right, font=\small, pos=0.85] {$a_3^{(1)}$} (7.2, -0.55);
\draw[trans] (s3) -- node[above, font=\small] {} (sT);

% Ellipsis for continuation
\node at (8.4, 0.3) {\small$\cdots$};

% Terminal rewards - redesigned layout
\node[reward, draw=blue!70, fill=blue!15, text=black] (r0) at (11.8, 0.55) {$R_0(\tau) = +1$};
\node[reward, draw=red!70, fill=red!15, text=black] (r1) at (11.8, -0.55) {$R_1(\tau) = -1$};

% Elegant curved arrows to rewards
\draw[->, >=stealth, blue!60, thick] (sT.east) -- ++(0.3, 0) |- (r0.west);
\draw[->, >=stealth, red!60, thick] (sT.east) -- ++(0.3, 0) |- (r1.west);

% Turn labels
\node[font=\normalsize, text=black!80] at (0, -0.85) {$t=0$};
\node[font=\normalsize, text=black!80] at (1.8, -0.85) {$t=1$};
\node[font=\normalsize, text=black!80] at (4.8, -0.85) {$t=2$};
\node[font=\normalsize, text=black!80] at (6.7, -0.85) {$t=3$};
\node[font=\normalsize, text=black!80] at (9.6, -0.85) {$t=T$};

% Legend for alternating turns
\node[font=\normalsize, text=blue!70] at (1.7, 1.5) {even turns};
\node[font=\normalsize, text=red!70] at (0.7, -1.5) {odd turns};

\end{tikzpicture}%
}
\caption{Two-player zero-sum Markov game structure. Both players share a single policy $\pi_\theta$ with role conditioning. Players alternate turns: Player~0 acts at even timesteps ($t \bmod 2 = 0$), Player~1 at odd timesteps. The transition function $T$ governs state dynamics based on actions. At terminal state $s_T$, rewards satisfy the zero-sum constraint $R_0(\tau) + R_1(\tau) = 0$.}
\label{fig:markov_game}
\end{figure}

For competitive interactions, we extend this to a two-player zero-sum Markov game \citep{littman1994markov} $\mathcal{G} = (\mathcal{S}, \mathcal{A}_0, \mathcal{A}_1, T, r, \gamma)$ with opposed rewards:
\begin{equation}
r_0 + r_1 = 0 \;\; \forall (s, a^{(0)}, a^{(1)}), \quad R_1(\tau) = -R_0(\tau).
\end{equation}
Figure~\ref{fig:markov_game} illustrates this structure for trajectory $\tau = \{(s_t, a_t^{(0)}, a_t^{(1)})\}_{t=0}^T$.

\subsection{SPIRAL}

SPIRAL \citep{liu2025spiral} trains language models through self-play on turn-based zero-sum games $\mathcal{G} = \{G_1, \ldots, G_n\}$ with sparse terminal rewards $R_p(\tau) \in \{-1, 0, 1\}$ (see Appendix~\ref{app:spiral} for details). Both players share a single policy $\pi_\theta$ with role conditioning: player $p = t \bmod 2$ generates $y_t^{(p)} \sim \pi_\theta(\cdot | s_t, p, G)$ at turn $t$.

To handle asymmetric expected returns across roles, SPIRAL employs Role-conditioned Advantage Estimation (RAE) with separate baselines $b_{G,p}$ per game-role pair:
\begin{align}
A_{G,p}(\tau) &= R_p(\tau) - b_{G,p}, \\
\nabla_\theta J &= \mathbb{E}\Big[ {\textstyle\sum_{t \in \mathcal{T}_p}} A_{G,p} \nabla_\theta \log \pi_\theta(y_t^{(p)} | s_t) \Big], \nonumber
\end{align}
where $\mathcal{T}_p$ indexes turns of player $p$ and baselines are updated via exponential moving average.
\begin{figure*}[t]
    \centering
    \resizebox{\textwidth}{!}{%
    \begin{tikzpicture}[
        % Node styles
        box/.style={rectangle, rounded corners=4pt, draw=black!70, minimum height=9mm, minimum width=20mm, font=\small, align=center},
        input/.style={box, fill=gray!10},
        spiral/.style={box, fill=orange!15, draw=orange!60},
        phi/.style={box, fill=blue!15, draw=blue!60},
        psi/.style={box, fill=green!15, draw=green!60},
        output/.style={box, fill=purple!15, draw=purple!60, minimum width=24mm},
        evaluator/.style={ellipse, draw=black!50, fill=yellow!10, minimum width=18mm, minimum height=7mm, font=\footnotesize},
        % Arrow styles
        arrow/.style={->, >=stealth, thick, black!70},
        multarrow/.style={->, >=stealth, thick, blue!60},
        addarrow/.style={->, >=stealth, thick, green!60},
    ]
    
    % Input trajectory
    \node[input] (tau) at (0, 0) {Trajectory $\tau$};
    
    % SPIRAL branch (top)
    \node[spiral] (agame) at (3.2, 0.8) {$A_{\text{game}}(\tau)$\\[-1pt]{\scriptsize Game Advantage}};

    % LLM Evaluator - same level as psi for horizontal arrow
    \node[evaluator] (llm) at (3.2, -0.8) {LLM $\mathcal{E}$};

    % TAM components
    \node[phi] (phi) at (6.0, 0.8) {$\varphi(\tau) \in \{0, 0.5, 1\}$\\[-1pt]{\scriptsize Transferability}};
    \node[psi] (psi) at (6.0, -0.8) {$\psi(\tau) \in \{-1, 0, +1\}$\\[-1pt]{\scriptsize Evolution}};

    % Multiplication and addition nodes
    \node[circle, draw=black!60, fill=white, inner sep=2pt, font=\small] (mult) at (8.6, 0.8) {$\times$};
    \node[circle, draw=black!60, fill=white, inner sep=2pt, font=\small] (add) at (9.8, 0) {$+$};

    % Beta coefficient
    \node[font=\footnotesize, text=green!60!black] at (8.6, -0.6) {$\beta$};
    
    % Output
    \node[output] (atam) at (12, 0) {$A_{\text{mod}}(\tau)$\\[-1pt]{\scriptsize Modulated Advantage}};
    
    % Arrows from trajectory
    \draw[arrow] (tau.east) -- ++(0.4, 0) |- (agame.west);
    \draw[arrow] (tau.east) -- ++(0.4, 0) |- (llm.west);
    
    % Arrow from SPIRAL to phi then multiplication
    \draw[arrow] (agame.east) -- (phi.west);
    
    % Arrows from LLM to psi - horizontal
    \draw[arrow] (llm.east) -- (psi.west);
    
    % Arrow from phi to multiplication
    \draw[multarrow] (phi.east) -- (mult.west);
    
    % Arrow from multiplication to addition
    \draw[arrow] (mult.east) -- (add.north);
    
    % Arrow from psi to addition
    \draw[addarrow] (psi.east) -| (add.south);
    
    % Arrow to output
    \draw[arrow] (add.east) -- (atam.west);
    
    % Light blue background for our contribution (TAM area)
    \begin{scope}[on background layer]
    \fill[blue!8, rounded corners=6pt] (4.4, -1.35) rectangle (10.6, 1.3);
    \end{scope}

    % Contribution label
    \node[font=\small, text=contributioncolor] at (7.9, 1.5) {\textbf{\method{}}};

    % Equation annotation - bottom right of the blue block
    \node[font=\scriptsize, text=black!45, anchor=south east] at (10.5, -1.3) {$A_{\text{mod}} = A_{\text{game}} \cdot \varphi + \beta \cdot \psi$};

    % Labels
    \node[font=\scriptsize, text=orange!60!black, above] at (1.6, 1.05) {SPIRAL};
    \node[font=\scriptsize, text=blue!60!black] at (4.9, 0.2) {\textsf{RTC}};
    \node[font=\scriptsize, text=green!60!black] at (4.9, -0.2) {\textsf{RER}};
    
    \end{tikzpicture}%
    }
    \caption{Overview of \method{}. Given a trajectory $\tau$ from self-play, the game-based advantage $A_{\text{game}}$ is computed. \method{} modulates this advantage using two signals: the Reasoning Transferability Coefficient $\varphi$ that multiplicatively scales the advantage based on cross-domain transfer potential, and the Reasoning Evolution Reward $\psi$ that additively rewards reasoning development within trajectories.}
    \label{fig:framework}
    \end{figure*}
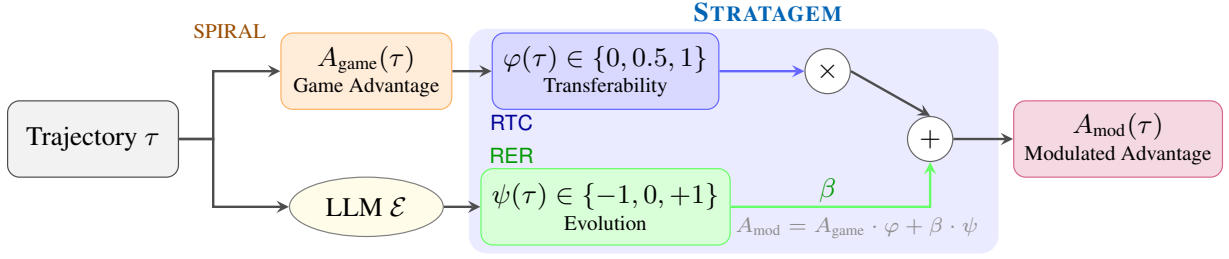

\section{Method}
\label{sec:method}

This section presents \method{}, which selectively reinforces transferable reasoning patterns through trajectory advantage modulation. We first provide an overview (\S\ref{sec:method_overview}), then detail the Reasoning Transferability Coefficient (\S\ref{sec:rtc}) and Reasoning Evolution Reward (\S\ref{sec:rer}).

\subsection{Overview}
\label{sec:method_overview}

Transferring reasoning capabilities from games to domains such as mathematics and coding faces two fundamental challenges:

\begin{enumerate}[leftmargin=*, topsep=2pt, itemsep=1pt]
\item \textbf{Domain Specificity}: Reasoning patterns learned from games tend to be anchored in game-specific concepts, terminology, and heuristics (e.g., ``King beats Queen'') rather than abstract, domain-agnostic patterns (e.g., ``enumerate cases and compute expected value'').

\item \textbf{Contextual Stasis}: Games present static problem contexts where the rules, setting, and problem description remain fixed throughout interaction. In contrast, mathematical problem-solving involves evolving contexts where decomposition creates new sub-problems, intermediate results reshape the solution space, and reasoning must continuously adapt to changing conditions.
\end{enumerate}

\noindent These challenges limit reasoning transfer: domain-specific patterns fail to generalize, and models trained on static contexts cannot adapt to evolving problem states. To incentivize transferable reasoning, we design \method{} to tackle both challenges through trajectory advantage modulation.

Given a trajectory $\tau$ from a zero-sum game, SPIRAL computes the role-conditioned advantage $A_{\text{game}}(\tau) = R_p(\tau) - b_{G,p}$ based solely on terminal game outcomes. \method{} extends this formulation by introducing two complementary signals designed to capture reasoning quality:
\begin{equation}
A_{\text{mod}}(\tau) = A_{\text{game}}(\tau) \cdot \varphi(\tau) + \beta \cdot \psi(\tau),
\label{eq:mod_advantage}
\end{equation}
\noindent where $\varphi(\tau) \in \{0, 0.5, 1\}$ is the \textbf{Reasoning Transferability Coefficient} that addresses \textit{domain specificity} by measuring the abstraction level of reasoning patterns (\S\ref{sec:rtc}), and $\psi(\tau) \in \{-1, 0, +1\}$ is the \textbf{Reasoning Evolution Reward} that addresses \textit{contextual stasis} by incentivizing reasoning that progressively adapts and deepens across turns (\S\ref{sec:rer}). The hyperparameter $\beta$ controls the relative contribution of the reasoning evolution.

This formulation achieves selective reinforcement through the multiplicative term $A_{\text{game}} \cdot \varphi$: trajectories with abstract, domain-agnostic reasoning ($\varphi \approx 1$) retain their full game-derived advantage, while those with domain-specific reasoning ($\varphi \approx 0$) have their influence diminished. The additive term $\beta \cdot \psi$ rewards trajectories that demonstrate progressive reasoning development, preparing the model for the evolving contexts of real-world problem-solving. Figure~\ref{fig:framework} illustrates this modulation framework.

\subsection{Reasoning Transferability Coefficient}
\label{sec:rtc}

\begin{figure}[t]
\centering
\resizebox{\columnwidth}{!}{%
\begin{tikzpicture}[
    % Node styles
    dim/.style={rectangle, rounded corners=3pt, minimum height=9mm, minimum width=28mm, font=\normalsize, align=center},
    alpha/.style={dim, draw=blue!60, fill=blue!10},
    sigma/.style={dim, draw=teal!60, fill=teal!10},
    rho/.style={dim, draw=violet!60, fill=violet!10},
    result/.style={rectangle, rounded corners=4pt, draw=blue!70, fill=blue!20, minimum height=14mm, minimum width=22mm, font=\large, align=center},
    weight/.style={circle, draw=black!50, fill=white, inner sep=2pt, font=\small},
]

% Three dimensions - more compact vertical spacing
\node[alpha] (alpha) at (0, 1.4) {\textbf{Abstraction} $\alpha$\\[-1pt]{\footnotesize Domain-agnostic}};
\node[sigma] (sigma) at (0, 0) {\textbf{Structure} $\sigma$\\[-1pt]{\footnotesize Reusable frameworks}};
\node[rho] (rho) at (0, -1.4) {\textbf{Principle} $\rho$\\[-1pt]{\footnotesize General principles}};

% Discrete scales [0, 0.5, 1] with tick marks and selected value markers
\foreach \y/\sel/\col in {1.4/2/blue, 0/1/teal, -1.4/2/violet} {
    % Scale line
    \draw[thick, gray!40] (2.2, \y) -- (4.8, \y);
    % Tick marks at 0, 0.5, 1
    \foreach \x/\lab in {2.2/0, 3.5/0.5, 4.8/1} {
        \draw[thick, gray!50] (\x, \y-0.15) -- (\x, \y+0.15);
        \node[font=\scriptsize, text=black!50] at (\x, \y-0.4) {$\lab$};
    }
    % Empty circles for unselected values
    \foreach \x in {2.2, 3.5, 4.8} {
        \draw[thick, \col!60] (\x, \y) circle (0.12);
    }
}
% Filled circles for selected values: α=1.0, σ=0.5, ρ=1.0
\fill[blue!70] (4.8, 1.4) circle (0.12);
\fill[teal!70] (3.5, 0) circle (0.12);
\fill[violet!70] (4.8, -1.4) circle (0.12);

% Weights
\node[weight] (w1) at (5.4, 1.4) {.35};
\node[weight] (w2) at (5.4, 0) {.35};
\node[weight] (w3) at (5.4, -1.4) {.30};

% Arrows to weights
\draw[->, >=stealth, thick, blue!50] (4.9, 1.4) -- (w1.west);
\draw[->, >=stealth, thick, teal!50] (4.9, 0) -- (w2.west);
\draw[->, >=stealth, thick, violet!50] (4.9, -1.4) -- (w3.west);

% Summation
\node[font=\Large] (sum) at (6.3, 0) {$\Sigma$};

% Arrows from weights to sum
\draw[->, >=stealth, thick, black!40] (w1.east) -- (sum.north west);
\draw[->, >=stealth, thick, black!40] (w2.east) -- (sum.west);
\draw[->, >=stealth, thick, black!40] (w3.east) -- (sum.south west);

% Result
\node[result] (phi) at (8.2, 0) {$\varphi(\tau)$\\[-1pt]{\small$\in [0, 1]$}};

% Arrow to result
\draw[->, >=stealth, thick, black!70] (sum.east) -- (phi.west);

% Formula at bottom
\node[font=\small, text=black!60] at (4, -2.3) {$\varphi = 0.35 \cdot \alpha + 0.35 \cdot \sigma + 0.30 \cdot \rho$};

\end{tikzpicture}%
}
\caption{Reasoning Transferability Coefficient $\varphi(\tau)$. Each dimension is scored discretely as $\{0, 0.5, 1\}$ (low/medium/high). The weighted sum quantifies cross-domain transfer potential.}
\label{fig:rtc}
\end{figure}
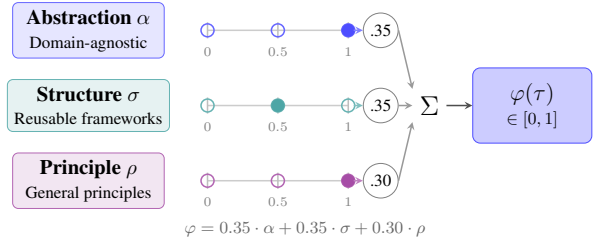

\paragraph{Motivation.}
The \textit{domain specificity} challenge arises because game training naturally produces reasoning tied to game semantics. Consider two reasoning traces from the same game:

\vspace{2pt}
\noindent
\begin{minipage}[t]{0.48\columnwidth}
\begin{tcolorbox}[
    colback=red!5, colframe=red!60!black,
    fonttitle=\footnotesize\bfseries,
    title={\textsf{Game-Specific} ($\varphi \approx 0$)},
    boxrule=0.5pt, arc=2pt, left=3pt, right=3pt, top=2pt, bottom=2pt
]
\footnotesize\itshape
``I have the lowest card and the opponent bet, which usually indicates strength. I should fold.''
\end{tcolorbox}
\end{minipage}%
\hfill
\begin{minipage}[t]{0.48\columnwidth}
\begin{tcolorbox}[
    colback=blue!5, colframe=blue!60!black,
    fonttitle=\footnotesize\bfseries,
    title={\textsf{Abstract} ($\varphi \approx 1$)},
    boxrule=0.5pt, arc=2pt, left=3pt, right=3pt, top=2pt, bottom=2pt
]
\footnotesize\itshape
``Enumerate cases: Case 1 yields $-2 \times 0.5 = -1$; Case 2 yields $+2 \times 0.5 = +1$. Select the option maximizing expected utility.''
\end{tcolorbox}
\end{minipage}
\vspace{2pt}

\noindent The first relies on game-specific heuristics with no utility outside its original context. The second employs case enumeration and expected value, frameworks applicable to any decision problem. To address domain specificity, we quantify how well reasoning patterns can transfer by measuring their abstraction level.

\paragraph{Formulation.}
We operationalize transferability through three dimensions that characterize domain-independent reasoning (Figure~\ref{fig:rtc}):

\begin{itemize}[leftmargin=*, nosep]
\item \textbf{Abstraction Level} ($\alpha$): The extent to which reasoning employs domain-agnostic concepts (e.g., ``expected value,'' ``probability distribution'') versus game-specific terminology (e.g., ``King beats Queen'').

\item \textbf{Structural Clarity} ($\sigma$): The presence of reusable reasoning frameworks such as case-by-case analysis, if-then chains, or systematic enumeration.

\item \textbf{Principle Orientation} ($\rho$): Whether reasoning invokes general principles (e.g., ``by Bayes' theorem,'' ``to maximize expected utility'') rather than experiential heuristics.
\end{itemize}

Each dimension is scored discretely as $\{0, 0.5, 1\}$ (low/medium/high) using a language model evaluator (prompt details in Appendix~\ref{app:rtc_prompt}). The transferability coefficient is:
\begin{equation}
\varphi(\tau) = w_\alpha \cdot \alpha(\tau) + w_\sigma \cdot \sigma(\tau) + w_\rho \cdot \rho(\tau),
\label{eq:rtc}
\end{equation}
where $w_\alpha = 0.35$, $w_\sigma = 0.35$, and $w_\rho = 0.30$ reflect the relative importance of each dimension.

\subsection{Reasoning Evolution Reward}
\label{sec:rer}

\begin{figure}[t]
\centering
\resizebox{\columnwidth}{!}{%
\begin{tikzpicture}[
    % Node styles
    dim/.style={rectangle, rounded corners=3pt, minimum height=9mm, minimum width=28mm, font=\normalsize, align=center},
    deep/.style={dim, draw=orange!60, fill=orange!10},
    adapt/.style={dim, draw=cyan!60, fill=cyan!10},
    coher/.style={dim, draw=red!60, fill=red!10},
    result/.style={rectangle, rounded corners=4pt, draw=green!70, fill=green!20, minimum height=14mm, minimum width=22mm, font=\large, align=center},
    weight/.style={circle, draw=black!50, fill=white, inner sep=2pt, font=\small},
]

% Three dimensions - more compact vertical spacing
\node[deep] (d) at (0, 1.4) {\textbf{Deepening} $d$\\[-1pt]{\footnotesize Simple $\to$ complex}};
\node[adapt] (a) at (0, 0) {\textbf{Adaptation} $a$\\[-1pt]{\footnotesize Adjusts to new info}};
\node[coher] (c) at (0, -1.4) {\textbf{Coherence} $c$\\[-1pt]{\footnotesize Logical consistency}};

% Discrete scales [-1, 0, +1] with tick marks and selected value markers
\foreach \y/\col in {1.4/orange, 0/cyan, -1.4/red} {
    % Scale line
    \draw[thick, gray!40] (2.2, \y) -- (4.8, \y);
    % Tick marks at -1, 0, +1
    \foreach \x/\lab in {2.2/-1, 3.5/0, 4.8/+1} {
        \draw[thick, gray!50] (\x, \y-0.15) -- (\x, \y+0.15);
        \node[font=\scriptsize, text=black!50] at (\x, \y-0.4) {$\lab$};
    }
    % Empty circles for unselected values
    \foreach \x in {2.2, 3.5, 4.8} {
        \draw[thick, \col!60] (\x, \y) circle (0.12);
    }
}
% Filled circles for selected values: d=+1, a=0, c=+1
\fill[orange!70] (4.8, 1.4) circle (0.12);
\fill[cyan!70] (3.5, 0) circle (0.12);
\fill[red!70] (4.8, -1.4) circle (0.12);

% Weights
\node[weight] (w1) at (5.4, 1.4) {.35};
\node[weight] (w2) at (5.4, 0) {.25};
\node[weight] (w3) at (5.4, -1.4) {.40};

% Arrows to weights
\draw[->, >=stealth, thick, orange!50] (4.9, 1.4) -- (w1.west);
\draw[->, >=stealth, thick, cyan!50] (4.9, 0) -- (w2.west);
\draw[->, >=stealth, thick, red!50] (4.9, -1.4) -- (w3.west);

% Summation
\node[font=\Large] (sum) at (6.3, 0) {$\Sigma$};

% Arrows from weights to sum
\draw[->, >=stealth, thick, black!40] (w1.east) -- (sum.north west);
\draw[->, >=stealth, thick, black!40] (w2.east) -- (sum.west);
\draw[->, >=stealth, thick, black!40] (w3.east) -- (sum.south west);

% Result
\node[result] (psi) at (8.2, 0) {$\psi(\tau)$\\[-1pt]{\small$\in [-1, 1]$}};

% Arrow to result
\draw[->, >=stealth, thick, black!70] (sum.east) -- (psi.west);

% Formula at bottom
\node[font=\small, text=black!60] at (4, -2.3) {$\psi = 0.35 \cdot d + 0.25 \cdot a + 0.40 \cdot c$};

\end{tikzpicture}%
}
\caption{Reasoning Evolution Reward $\psi(\tau)$. Each dimension is scored as $\{-1, 0, +1\}$ (degradation/neutral/improvement). The zero-centered design reduces variance while penalizing degradation.}
\label{fig:rer}
\end{figure}
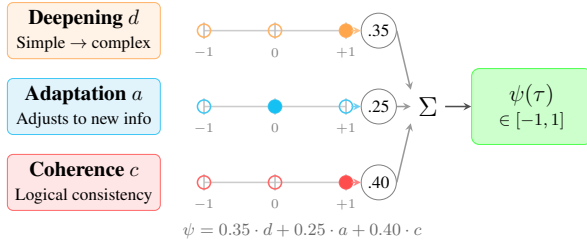

\paragraph{Motivation.}
The \textit{contextual stasis} challenge stems from the static nature of game environments: rules remain fixed, and shallow pattern-matching suffices for winning. Solving a math problem, by contrast, requires continuously evolving reasoning where each step reshapes the solution space. Consider two multi-turn reasoning traces:

\vspace{2pt}
\noindent
\begin{minipage}[t]{0.48\columnwidth}
\begin{tcolorbox}[
    colback=red!5, colframe=red!60!black,
    fonttitle=\footnotesize\bfseries,
    title={\textsf{Static} ($\psi \approx -1$)},
    boxrule=0.5pt, arc=2pt, left=3pt, right=3pt, top=2pt, bottom=2pt
]
\footnotesize\itshape
``T1: I see the board state. T2: The opponent moved. T3: I will respond with my usual strategy.''
\end{tcolorbox}
\end{minipage}%
\hfill
\begin{minipage}[t]{0.48\columnwidth}
\begin{tcolorbox}[
    colback=blue!5, colframe=blue!60!black,
    fonttitle=\footnotesize\bfseries,
    title={\textsf{Evolving} ($\psi \approx 1$)},
    boxrule=0.5pt, arc=2pt, left=3pt, right=3pt, top=2pt, bottom=2pt
]
\footnotesize\itshape
``T1: Center opening signals control. T2: Corner response confirms defensive pattern. T3: Exploit via diagonal trap.''
\end{tcolorbox}
\end{minipage}
\vspace{2pt}

\noindent The first exhibits shallow, repetitive observations without adaptation. The second progressively deepens analysis, adapts to opponent behavior, and builds coherently on prior conclusions. To address contextual stasis, we introduce a reward signal that explicitly encourages such reasoning evolution within trajectories.

\paragraph{Formulation.}
The reasoning evolution reward captures three aspects of within-trajectory reasoning dynamics (Figure~\ref{fig:rer}):

\begin{itemize}[leftmargin=*, nosep]
\item \textbf{Reasoning Deepening} ($d$): Whether reasoning progresses from simple observations to complex analysis across turns, analogous to building mathematical proofs incrementally.

\item \textbf{Strategy Adaptation} ($a$): The degree to which reasoning adjusts based on observed opponent behavior or evolving game states, reflecting the ability to incorporate new information.

\item \textbf{Logical Coherence} ($c$): Whether later reasoning builds on earlier conclusions, maintaining a consistent logical thread throughout the trajectory.
\end{itemize}

Each dimension is scored discretely as $\{-1, 0, +1\}$: $+1$ indicates improvement, $0$ indicates neutral performance, and $-1$ indicates degradation. The zero-centered design aligns naturally with the advantage function. The evolution reward is:
\begin{equation}
\psi(\tau) = w_d \cdot d(\tau) + w_a \cdot a(\tau) + w_c \cdot c(\tau),
\label{eq:rer}
\end{equation}
where $w_c = 0.40$, $w_d = 0.35$, and $w_a = 0.25$ prioritize logical coherence as the foundation of sound reasoning. Evaluation prompts are provided in Appendix~\ref{app:rer_prompt}.

\paragraph{Design Rationale.}
The choice of $\psi \in [-1, 1]$ serves two purposes. First, zero-centering reduces variance in policy gradient estimates since the expected value of $\psi$ centers around zero rather than a positive constant. Second, negative values actively discourage reasoning degradation: trajectories where reasoning quality deteriorates receive reduced reinforcement even if they achieve favorable game outcomes.

\subsection{Training Procedure}

\begin{figure}[t]
\centering
\begin{tcolorbox}[
    colback=gray!3,
    colframe=black!60,
    boxrule=0.4pt,
    arc=2mm,
    title={\textbf{Algorithm 1:} \method{} Training},
    fonttitle=\small,
    coltitle=black,
    colbacktitle=gray!15,
    left=3pt, right=3pt, top=2pt, bottom=2pt
]
\small
\textbf{Input:} Policy $\pi_\theta$, game set $\mathcal{G}$, evaluator $\mathcal{E}$, coefficients $\beta$, $\eta$, $\alpha$\\[2pt]
\textbf{Output:} Trained policy $\pi_\theta$\\[4pt]
\textbf{for} iteration $= 1, 2, \ldots$ \textbf{do}

\begin{tcolorbox}[
    colback=orange!8,
    colframe=orange!50!black,
    boxrule=0.3pt,
    arc=1mm,
    left=2pt, right=2pt, top=1pt, bottom=1pt
]
\textcolor{orange!70!black}{\textit{// Step 1: Self-Play Trajectory Generation}}\\
$G \sim \mathcal{G}$, \, $p \sim \{0, 1\}$ \hfill \textcolor{gray}{\scriptsize $\triangleright$ Sample game and role}\\
$\tau = \{(s_t, y_t^{(p)})\} \leftarrow$ \textcolor{blue!70!black}{\texttt{SelfPlay}}$(\pi_\theta, G, p)$\\
$R_p(\tau) \in \{-1, 0, +1\} \leftarrow$ \textcolor{blue!70!black}{\texttt{GameOutcome}}$(\tau)$
\end{tcolorbox}

\begin{tcolorbox}[
    colback=teal!8,
    colframe=teal!50!black,
    boxrule=0.3pt,
    arc=1mm,
    left=2pt, right=2pt, top=1pt, bottom=1pt
]
\textcolor{teal!70!black}{\textit{// Step 2: Game-Based Advantage (SPIRAL)}}\\
$A_{\text{game}}(\tau) \leftarrow R_p(\tau) - b_{G,p}$ \hfill \textcolor{gray}{\scriptsize $\triangleright$ Role-conditioned advantage}\\
$b_{G,p} \leftarrow \eta \cdot R_p(\tau) + (1-\eta) \cdot b_{G,p}$ \hfill \textcolor{gray}{\scriptsize $\triangleright$ EMA baseline update}
\end{tcolorbox}

\begin{tcolorbox}[
    colback=blue!10,
    colframe=contributioncolor,
    boxrule=0.5pt,
    arc=1mm,
    left=2pt, right=2pt, top=1pt, bottom=1pt
]
\textcolor{contributioncolor}{\textit{// Step 3: \textbf{\method{} Modulation (Ours)}}}\\
$\varphi(\tau) \leftarrow$ \textcolor{blue!70!black}{\texttt{EvalRTC}}$(\mathcal{E}, \tau)$ \hfill \textcolor{contributioncolor}{\scriptsize $\triangleright$ \colorbox{yellow!30}{Transferability} (\S\ref{sec:rtc})}\\
$\psi(\tau) \leftarrow$ \textcolor{blue!70!black}{\texttt{EvalRER}}$(\mathcal{E}, \tau)$ \hfill \textcolor{contributioncolor}{\scriptsize $\triangleright$ \colorbox{yellow!30}{Evolution} (\S\ref{sec:rer})}\\
$A_{\text{mod}}(\tau) \leftarrow A_{\text{game}}(\tau) \cdot \varphi(\tau) + \beta \cdot \psi(\tau)$
\end{tcolorbox}

\begin{tcolorbox}[
    colback=green!8,
    colframe=green!50!black,
    boxrule=0.3pt,
    arc=1mm,
    left=2pt, right=2pt, top=1pt, bottom=1pt
]
\textcolor{green!60!black}{\textit{// Step 4: Policy Gradient Update}}\\
$\nabla_\theta J \leftarrow \sum_{t \in \mathcal{T}_p} A_{\text{mod}}(\tau) \nabla_\theta \log \pi_\theta(y_t^{(p)} \mid s_t)$\\
$\theta \leftarrow \theta + \alpha \nabla_\theta J$
\end{tcolorbox}

\textbf{end for}
\end{tcolorbox}
\caption{\method{} training procedure. Step 3 (blue box) highlights our contribution: trajectory advantage modulation incorporates transferability ($\varphi$) and evolution ($\psi$) signals.}
\label{alg:method}
\end{figure}

The training procedure (Figure~\ref{alg:method}) extends self-play with trajectory advantage modulation, where Step 3 constitutes our contribution.

\paragraph{Computational Considerations.}
To manage evaluation cost, we employ trajectory sampling where only a fraction undergo full LLM evaluation, with others assigned the batch mean.

\paragraph{Synergy Between Components.}
The components work together: $\varphi$ addresses \textit{domain specificity} via abstract pattern identification, while $\psi$ addresses \textit{contextual stasis} via adaptive reasoning rewards. Only trajectories exhibiting both qualities receive maximal reinforcement.

\section{Experiment}
\label{sec:experiment}

This section describes our experimental setup for evaluating \method{}. We introduce the game environments (\S\ref{sec:game_envs}), training configuration (\S\ref{sec:training_settings}), and evaluation metrics (\S\ref{sec:eval_metrics}).

\subsection{Game Environments}
\label{sec:game_envs}

Following \citet{liu2025spiral}, we adopt three text-based zero-sum games from TextArena~\citep{guertler2025textarena}: Tic-Tac-Toe for \textit{spatial reasoning}, Kuhn Poker~\citep{kuhn1950simplified} for \textit{probabilistic reasoning}, and Simple Negotiation for \textit{strategic optimization}. These games provide complementary coverage of core reasoning dimensions while offering naturally verifiable rewards through win/loss outcomes. Detailed game descriptions are provided in Appendix~\ref{app:game_envs}.

\subsection{Training Settings}
\label{sec:training_settings}

We build upon SPIRAL~\citep{liu2025spiral} using Qwen3-4B-Base~\citep{yang2025qwen3} as the base model. For trajectory advantage modulation, we set $\beta = 0.2$ and compute $\varphi$ and $\psi$ using GPT-4 as the evaluation backbone. With trajectory subsampling, GPT-4 scoring adds roughly \$100 per training run, negligible relative to the $\sim$30 GPU-hours on 2$\times$A100 required for training. Training runs on 2 NVIDIA A100 GPUs with vLLM~\citep{kwon2023efficient} for efficient inference. Complete hyperparameters and prompts are provided in Appendix~\ref{app:training_settings}.

\subsection{Evaluation Metrics}
\label{sec:eval_metrics}

We evaluate reasoning transfer across three categories: (1) \textit{mathematical reasoning} using MATH500~\citep{hendrycksmath2021}, OlympiadBench~\citep{he2024olympiadbench}, Minerva Math~\citep{lewkowycz2022solving}, AIME'24, AIME'25, and AMC'23; (2) \textit{general reasoning} using GPQA~\citep{rein2024gpqa} and MMLU-Pro~\citep{wang2024mmlu}; and (3) \textit{code generation} using HumanEval~\citep{chen2021evaluating} (pass@1). All evaluations use zero-shot prompting with prompts in Appendix~\ref{app:eval_prompts}.

\section{Results}
\label{sec:results}

\subsection{Main Results}
\label{sec:main_results}

\begin{figure*}[t]
\centering
\includegraphics[width=0.95\textwidth]{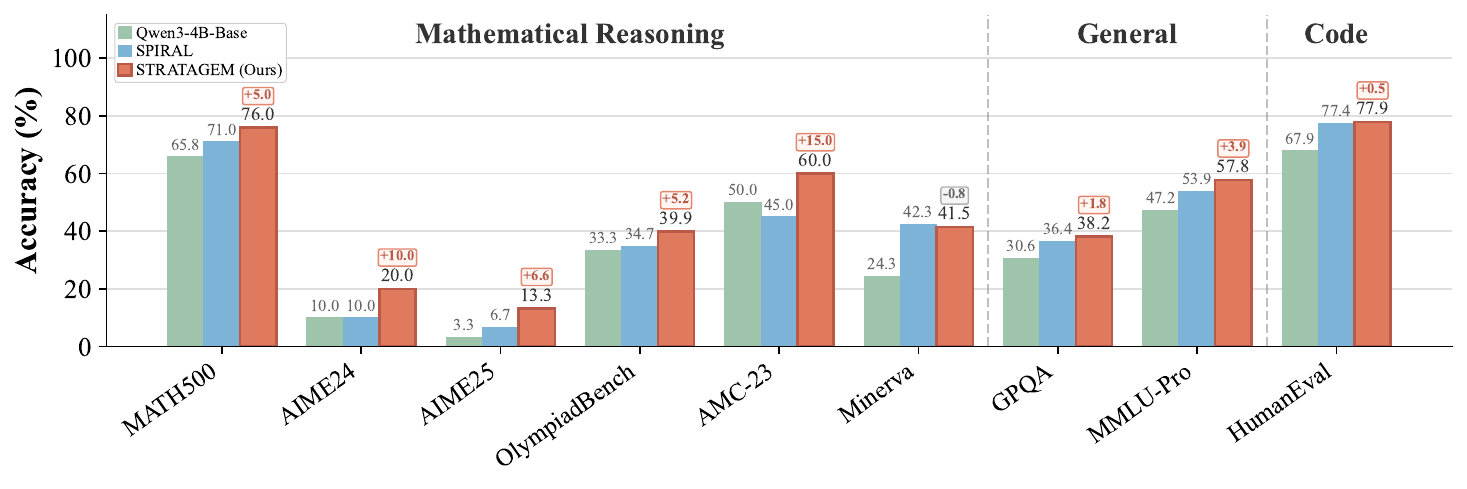}
\caption{Performance comparison across mathematical reasoning, general reasoning, and code generation benchmarks. \method{} consistently outperforms both Qwen3-4B-Base and SPIRAL, with particularly strong gains on competition-level mathematical tasks.}
\label{fig:main_results}
\end{figure*}

\begin{figure*}[t]
    \centering
    \includegraphics[width=0.95\textwidth]{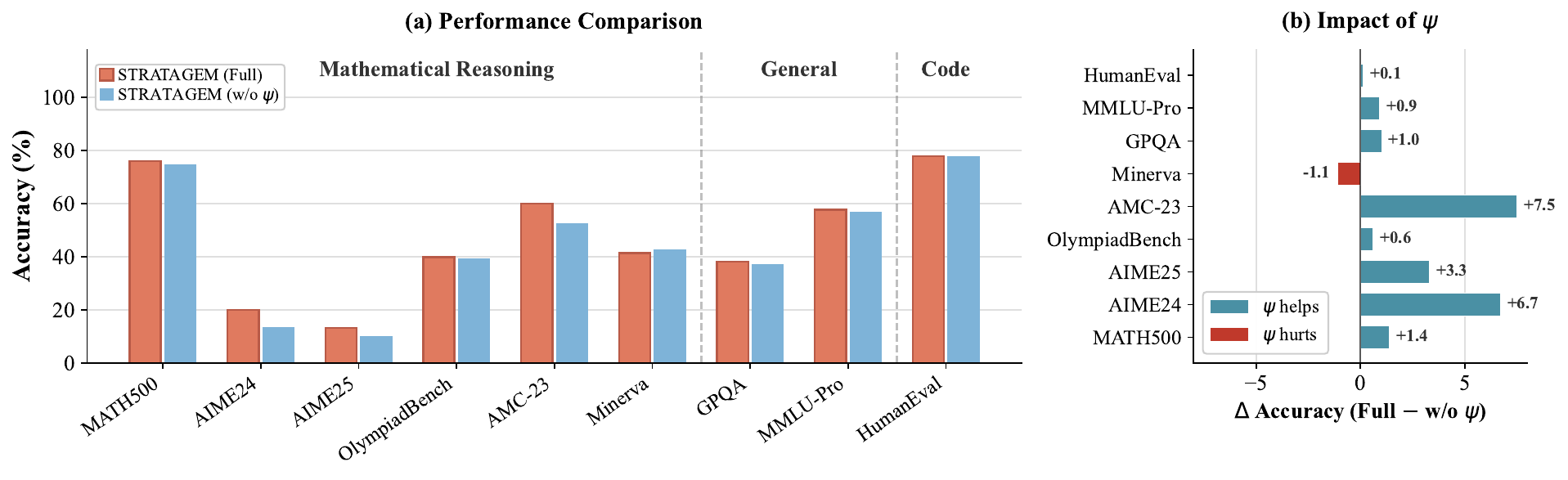}
    \caption{Ablation study on the Reasoning Evolution Reward ($\psi$). (a) Performance comparison between full \method{} and the variant without $\psi$. (b) Impact analysis showing $\psi$'s contribution across benchmarks.}
    \label{fig:ablation}
    \end{figure*}

Figure~\ref{fig:main_results} presents benchmark comparisons (details in Appendix~\ref{app:detailed_results}). \method{} achieves consistent improvements, with substantial gains on competition-level mathematics: AIME24 doubles (10\%$\rightarrow$20\%), AIME25 improves 4$\times$ (3.3\%$\rightarrow$13.3\%), and AMC-23 reaches 60\% versus baseline (50\%) and SPIRAL\footnote{SPIRAL results obtained using official codebase under identical configuration.} (45\%). MATH500 achieves 76\% (+5 over SPIRAL). Transfer extends to general reasoning (GPQA: 38.23\%, MMLU-Pro: 57.83\%) and code generation (HumanEval: 77.93\%, +10 over baseline), confirming that addressing \textit{domain specificity} ($\varphi$) and \textit{contextual stasis} ($\psi$) promotes transferable reasoning.

\subsection{Ablation Study}
\label{sec:ablation}

To isolate component contributions, we ablate $\psi$ (Figure~\ref{fig:ablation}; details in Appendix~\ref{app:ablation}). Removing $\psi$ causes substantial degradation on competition-level mathematics: AIME24 drops 6.70\% and AMC-23 drops 7.50\%, benchmarks demanding extended multi-step reasoning. Overall, $\psi$ improves 8 of 9 benchmarks, with consistent gains on general reasoning and code generation. Both components address complementary challenges: $\varphi$ ensures abstract reasoning (\textit{domain specificity}), while $\psi$ rewards adaptive reasoning (\textit{contextual stasis}). Both are necessary for robust transfer.

\subsection{Parameter Sensitivity}
\label{sec:param_sensitivity}

\begin{figure*}[t]
\centering
\includegraphics[width=0.95\textwidth]{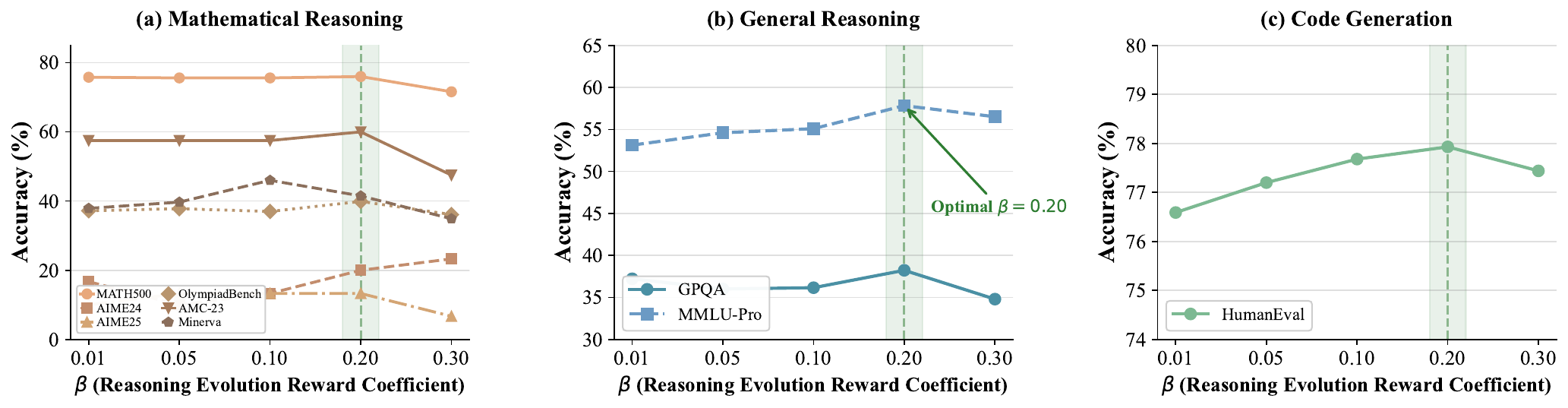}
\caption{Parameter sensitivity analysis for $\beta$. The green shaded region indicates the optimal value $\beta = 0.20$.}
\label{fig:beta_sensitivity}
\end{figure*}

The coefficient $\beta$ (Equation~\ref{eq:mod_advantage}) controls the balance between game-based advantage and reasoning evolution (Figure~\ref{fig:beta_sensitivity}; details in Appendix~\ref{app:beta_sensitivity}). Optimal performance occurs at $\beta = 0.20$, achieving peak scores on most benchmarks. Both extremes degrade performance: $\beta = 0.01$ contributes minimally, while $\beta = 0.30$ destabilizes training. Notably, high-complexity problems (AIME24) benefit from stronger $\beta$, while knowledge-focused tasks (Minerva) prefer weaker values.

\subsection{Human Evaluation}
\label{sec:human_eval}

\begin{figure}[t]
\centering
\includegraphics[width=\columnwidth]{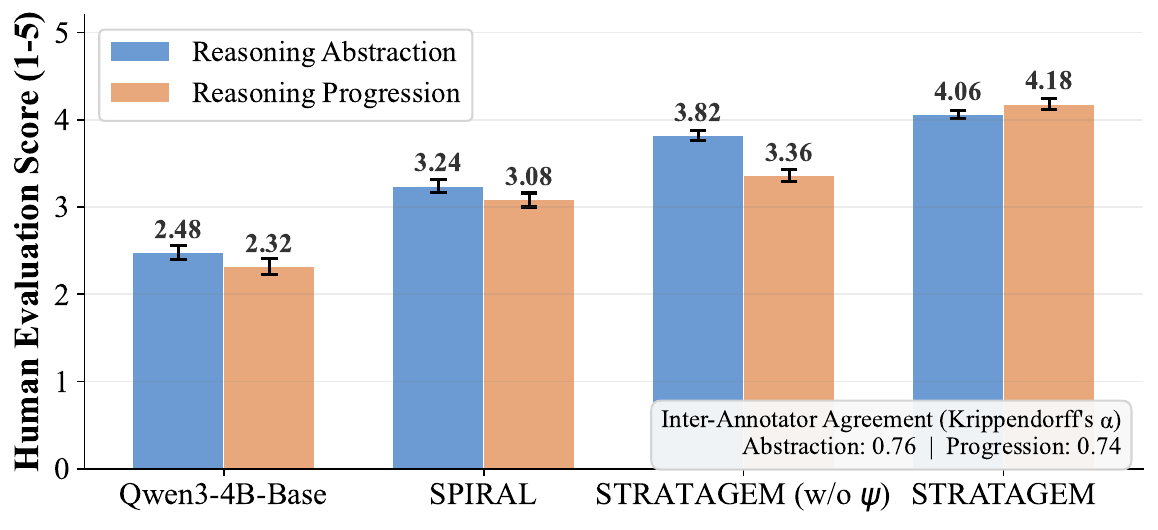}
\caption{Human evaluation results across two dimensions. Error bars indicate standard error. \method{} achieves the highest scores on both dimensions, while the ablated variant (w/o $\psi$) shows strong abstraction but weaker progression.}
\label{fig:human_eval}
\end{figure}

To complement automatic benchmarks, we conduct human evaluation on reasoning quality (Figure~\ref{fig:human_eval}). Five expert annotators evaluate 50 randomly sampled game trajectories along two dimensions on a 1 to 5 Likert scale: \textbf{Reasoning Abstraction} (domain-agnostic concepts vs.\ game-specific heuristics, corresponding to $\varphi$) and \textbf{Reasoning Progression} (deepening and coherence across steps, corresponding to $\psi$). \method{} achieves the highest scores on both dimensions (Abstraction: 4.06, Progression: 4.18), significantly outperforming baseline (2.48, 2.32) and SPIRAL (3.24, 3.08). The ablated variant without $\psi$ achieves competitive abstraction (3.82) but lower progression (3.36), confirming that $\psi$ specifically enhances reasoning evolution. Inter-annotator agreement is strong (Krippendorff's $\alpha \approx 0.75$). Guidelines are provided in Appendix~\ref{app:human_eval}.

\subsection{Evaluator Robustness}
\label{sec:evaluator_robustness}

\begin{table}[t]
\centering
\small
\resizebox{\columnwidth}{!}{
\begin{tabular}{l|ccc}
\toprule
\textbf{Agreement} & \textbf{GPT-4 vs.\ Claude} & \textbf{GPT-4 vs.\ Gemini} & \textbf{Claude vs.\ Gemini} \\
\midrule
Cohen's $\kappa$ ($\varphi$) & 0.71 & 0.67 & 0.64 \\
Cohen's $\kappa$ ($\psi$)    & 0.68 & 0.63 & 0.61 \\
Spearman $\rho$ ($\varphi$)  & 0.82 & 0.77 & 0.74 \\
Spearman $\rho$ ($\psi$)     & 0.79 & 0.74 & 0.71 \\
\bottomrule
\end{tabular}}
\caption{Cross-evaluator agreement on $\sim$200 trajectories re-scored with GPT-4, Claude 3.5 Sonnet, and Gemini 2.0 Flash. All $\kappa$ values exceed 0.60 (substantial agreement) and all Spearman correlations exceed 0.70, indicating that $\varphi$/$\psi$ scoring tracks objective trajectory properties rather than evaluator-specific biases.}
\label{tab:evaluator_consistency}
\end{table}

A natural concern with using GPT-4 as the $\varphi$/$\psi$ scorer is whether the reward signal reflects evaluator-specific preferences rather than intrinsic trajectory quality. To test this, we re-score $\sim$200 sampled trajectories with Claude 3.5 Sonnet and Gemini 2.0 Flash and measure pairwise agreement with GPT-4 (Table~\ref{tab:evaluator_consistency}). All Cohen's $\kappa$ values exceed 0.60 and all Spearman correlations exceed 0.70, placing agreement in the substantial-to-strong range across both dimensions and every evaluator pair. Combined with the human evaluation result (Krippendorff's $\alpha \approx 0.75$, \S\ref{sec:human_eval}), this indicates that $\varphi$ and $\psi$ capture properties recognizable across models and human experts rather than GPT-4 idiosyncrasies. Full scoring prompts are provided in Figures~\ref{fig:rtc_prompt} and~\ref{fig:rer_prompt}, enabling exact reproduction.

\subsection{Training Dynamics}
\label{sec:training_dynamics}

\begin{figure}[t]
    \centering
    \includegraphics[width=\columnwidth]{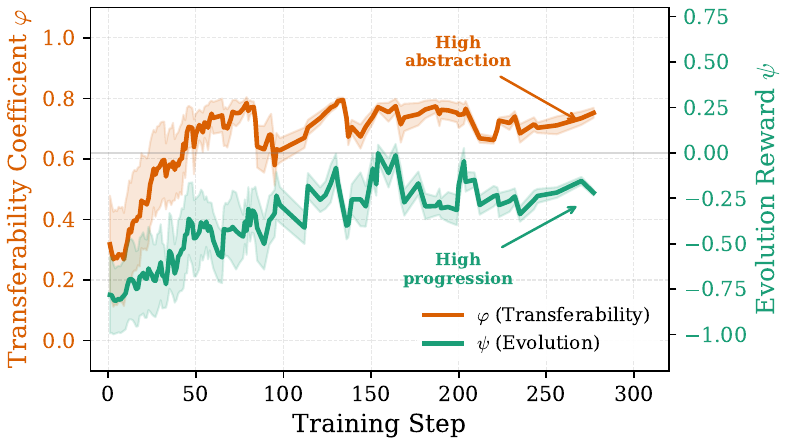}
    \caption{Evolution of \method{}'s modulation components during training. Both $\varphi$ (transferability) and $\psi$ (evolution) increase as training progresses, indicating the model learns abstract reasoning patterns and progressive reasoning chains.}
    \label{fig:tam_dynamics}
    \end{figure}

Figure~\ref{fig:tam_dynamics} reveals how \method{}'s modulation components evolve during training. The transferability coefficient $\varphi$ starts low, reflecting initial reliance on game-specific patterns, then steadily increases to 0.7 to 0.8 as the model learns abstract reasoning. The evolution reward $\psi$ follows a similar trend: initially negative (fragmented reasoning), it rises toward positive territory as coherent, progressive reasoning develops. These dynamics confirm that \method{} successfully guides training toward both abstraction and progression.

\subsection{Case Study: Reasoning Quality}
\label{sec:case_study}

\begin{figure*}[t]
\centering
\begin{tikzpicture}[
    board/.style={draw, minimum size=0.55cm, font=\footnotesize\bfseries},
    box/.style={draw, rounded corners=2pt, inner sep=4pt, font=\scriptsize, align=left, text width=5.2cm},
    baselinebox/.style={box, draw=baselinecolor, fill=baselinecolor!8},
    methodbox/.style={box, draw=progresscolor, fill=progresscolor!8}
]

% Baseline Box (Left)
\node[baselinebox, anchor=east] (baseline) at (-1.2, 1.4) {
\textbf{Baseline (Turn 2):}\\[2pt]
``On my \textcolor{baselinecolor}{\textbf{first move}}... take a \textcolor{baselinecolor}{\textbf{central position}} to maximize control.''\\[2pt]
\textcolor{baselinecolor}{\ding{56}} \textit{Repeats ``first move'' every turn}\\
\textcolor{baselinecolor}{\ding{56}} \textit{No awareness of prior moves}
};

% Tic-Tac-Toe Board (Center)
\node[font=\scriptsize\bfseries, anchor=south] at (0, 2.15) {Turn 2: O to move};
\draw[thick] (-0.3, 0.6) -- (-0.3, 2.1);
\draw[thick] (0.3, 0.6) -- (0.3, 2.1);
\draw[thick] (-0.9, 1.1) -- (0.9, 1.1);
\draw[thick] (-0.9, 1.6) -- (0.9, 1.6);
\node[board] at (-0.6, 1.85) {0};
\node[board, text=baselinecolor] at (0, 1.85) {X};
\node[board] at (0.6, 1.85) {2};
\node[board] at (-0.6, 1.35) {3};
\node[board, text=progresscolor] at (0, 1.35) {O};
\node[board] at (0.6, 1.35) {5};
\node[board] at (-0.6, 0.85) {6};
\node[board] at (0, 0.85) {7};
\node[board] at (0.6, 0.85) {8};

% \method{} Box (Right)
\node[methodbox, anchor=west] (gamerl) at (1.2, 1.4) {
\textbf{\method{} (Turn 2):}\\[2pt]
``Player 0 \textcolor{progresscolor}{\textbf{already has the center}}.''\\[1pt]
``\textcolor{abstractcolor}{\textbf{Threat Minimization}}: [1] limits options and sets up \textcolor{abstractcolor}{\textbf{future threats}}.''\\[2pt]
\textcolor{progresscolor}{\ding{52}} \textit{Recognizes prior state}\\
\textcolor{abstractcolor}{\ding{52}} \textit{Abstract strategic concepts}
};

% Legend at bottom (centered)
\node[font=\scriptsize, anchor=north] at (0, 0.35) {
\textcolor{abstractcolor}{\rule{0.5em}{0.5em}} Abstraction ($\varphi$) \quad
\textcolor{progresscolor}{\rule{0.5em}{0.5em}} Progression ($\psi$) \quad
\textcolor{baselinecolor}{\rule{0.5em}{0.5em}} Baseline issue
};

\end{tikzpicture}
\caption{Case study comparing reasoning traces on Tic-Tac-Toe. The baseline exhibits a ``reset issue'', repeating ``first move'' regardless of game state. \method{} demonstrates both \textcolor{abstractcolor}{abstraction} (strategic concepts) and \textcolor{progresscolor}{progression} (state awareness), corresponding to behaviors incentivized by $\varphi$ and $\psi$.}
\label{fig:case_study}
\end{figure*}
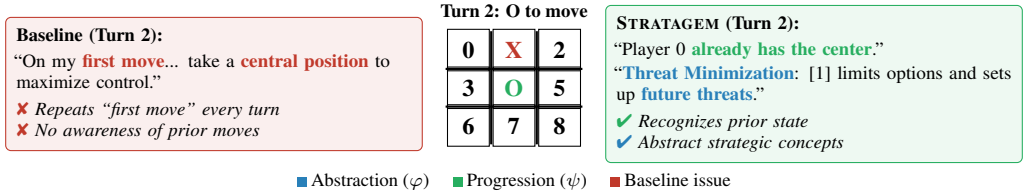

Figure~\ref{fig:case_study} compares reasoning traces from Tic-Tac-Toe (additional cases in Appendix~\ref{app:case_study}). The baseline exhibits a ``reset issue'': it generates reasoning as if every turn were the first, failing to track game state, a manifestation of \textit{contextual stasis}. It also relies on generic templates rather than adaptive strategies, reflecting \textit{domain specificity}. In contrast, \method{} demonstrates both properties our method cultivates. For \textit{abstraction}, it employs domain-agnostic concepts like ``Threat Minimization'' that transfer beyond specific board positions, patterns encouraged by $\varphi$. For \textit{progression}, it maintains state awareness (``already has the center'') and adapts strategy accordingly, behaviors incentivized by $\psi$. These complementary properties produce the structured decomposition and adaptive analysis essential for mathematical problem-solving.

\subsection{Out-of-Distribution Game Generalization}
\label{sec:ood_games}

\begin{table}[t]
\centering
\resizebox{\columnwidth}{!}{
\begin{tabular}{l|ccc}
\toprule
\textbf{Method} & \textbf{Snake} & \textbf{Pig Dice} & \textbf{Truth \& Deception} \\
\midrule
SPIRAL & 0.15 & 0.76 & 0.72 \\
\textbf{\method{}} & \textbf{0.35} & \textbf{0.96} & \textbf{0.80} \\
\midrule
$\Delta$ & \textcolor{blue}{+0.20} & \textcolor{blue}{+0.20} & \textcolor{blue}{+0.08} \\
\bottomrule
\end{tabular}}
\caption{Win rates against Gemini-2.0-Flash on out-of-distribution games (10 matches per game, randomized starting player). Game descriptions in Appendix~\ref{app:ood_games}.}
\label{tab:ood_games}
\end{table}

\begin{table*}[t]
    \centering
    \small
    \begin{tabular}{l|cccccc|cc|c}
    \toprule
    & \multicolumn{6}{c|}{\textbf{Mathematical Reasoning}} & \multicolumn{2}{c|}{\textbf{General}} & \textbf{Code} \\
    \cmidrule(lr){2-7} \cmidrule(lr){8-9} \cmidrule(lr){10-10}
    \textbf{Training Game} & MATH & AIME & AIME & Olympiad & AMC & Minerva & GPQA & MMLU & Human \\
    & 500 & 24 & 25 & Bench & 23 & Math & & Pro & Eval \\
    \midrule
    Tic-Tac-Toe & \underline{76.40} & 13.30 & 13.30 & 38.40 & 52.50 & 38.20 & 36.87 & 56.68 & \cellcolor{highlightblue}\textbf{78.54} \\
    Kuhn Poker & \cellcolor{highlightblue}\textbf{76.60} & 13.30 & 13.30 & \underline{39.40} & \underline{57.50} & \underline{41.20} & \underline{37.22} & \underline{57.14} & 77.32 \\
    Simple Negotiation & 73.60 & 10.00 & 13.30 & 37.50 & 52.50 & \cellcolor{highlightblue}\textbf{42.30} & 37.27 & 56.82 & \underline{78.17} \\
    \midrule
    \rowcolor{highlightblue!50}
    \textbf{\method{} (All Games)} & 76.00 & \textbf{20.00} & \textbf{13.30} & \textbf{39.90} & \textbf{60.00} & 41.50 & \textbf{38.23} & \textbf{57.83} & 77.93 \\
    \bottomrule
    \end{tabular}
    \caption{Single-game vs multi-game training comparison. \textbf{Bold} with \colorbox{highlightblue}{blue background} indicates best performance; \underline{underline} indicates second best. Multi-game training achieves best results on 6/9 benchmarks, with particularly strong gains on competition-level mathematics (AIME24, AMC-23).}
    \label{tab:single_game}
    \end{table*}

Following \citet{liu2025spiral}, we evaluate generalization to unseen games (Table~\ref{tab:ood_games}). \method{} outperforms SPIRAL across three OOD games: Snake ($+$0.20), Pig Dice ($+$0.20), and Truth and Deception ($+$0.08). These gains confirm that $\varphi$ and $\psi$ cultivate reasoning patterns rather than game-specific heuristics, enabling robust performance on novel challenges.

\subsection{Single-Game vs Multi-Game Training}
\label{sec:single_game}

To assess whether game diversity aids transfer, we compare single-game versus multi-game training (Table~\ref{tab:single_game}). Multi-game training achieves best performance on 6 of 9 benchmarks, with pronounced gains on competition-level mathematics (AIME24: $+$6.70\%, AMC-23: $+$2.50\%). While single-game training excels on benchmarks reflecting skill-task alignments, multi-game training produces robust generalization by combining reasoning patterns, particularly for complex problems.

\subsection{Generalization to Instruction-Tuned Models}
\label{sec:instruct_generalization}

\begin{table}[t]
\centering
\small
\begin{tabular}{l|cc|c}
\toprule
\textbf{Benchmark} & \textbf{SPIRAL} & \textbf{\method{}} & $\Delta$ \\
\midrule
MATH500   & 79.40 & \textbf{82.20} & \textcolor{blue}{+2.80} \\
AIME24    & 16.70 & \textbf{23.30} & \textcolor{blue}{+6.60} \\
AMC-23    & 57.50 & \textbf{65.00} & \textcolor{blue}{+7.50} \\
GPQA      & 40.18 & \textbf{41.74} & \textcolor{blue}{+1.56} \\
HumanEval & 81.10 & \textbf{81.40} & \textcolor{blue}{+0.30} \\
\bottomrule
\end{tabular}
\caption{\method{} vs.\ SPIRAL on Qwen3-4B-Instruct. \method{} provides consistent gains across mathematical, general, and code benchmarks, confirming that trajectory-advantage modulation is not tied to base-model initialization.}
\label{tab:qwen_instruct}
\end{table}

Beyond generalization across games, a natural question is whether \method{} is tied to a base-model initialization. We therefore apply the same training pipeline to Qwen3-4B-Instruct, a model that already exhibits instruction-following behavior, and compare against SPIRAL under identical settings (Table~\ref{tab:qwen_instruct}). \method{} improves over SPIRAL on all five benchmarks, with the largest gains on competition-level mathematics (AIME24: $+$6.60, AMC-23: $+$7.50), while general reasoning (GPQA: $+$1.56) and code generation (HumanEval: $+$0.30) also improve. The smaller absolute deltas relative to the base-model setting reflect reduced headroom rather than diminished effect, since trajectory-advantage modulation operates on the reward signal and is therefore architecture- and initialization-agnostic. These results indicate that $\varphi$ and $\psi$ capture reasoning properties that remain useful even when the starting policy has already been aligned.

\section{Related Work}

Games have served as fundamental AI testbeds, with systems like AlphaGo \citep{silver2016mastering}, OpenAI Five \citep{berner2019dota}, and AlphaStar \citep{vinyals2019grandmaster} achieving superhuman performance through self-play. This paradigm has been extended to LLM-based game agents across strategic games \citep{meta2022human,xu2023exploring,qi2024civrealm,feng2024survey}, text-based arenas \citep{guertler2025textarena,hudi2025textgames}, and comprehensive benchmarks \citep{Park2025OrakAF,hu2025lmgame,cipolina2025game,guo2026game}. Reinforcement learning has emerged as a powerful approach for LLM reasoning \citep{guo2025deepseek,zhang2025survey,zhao2025absolute,Chen2025TowardsRE,feng2025reasoning}, with self-play adapted through adversarial games \citep{cheng2024self}, theorem proving \citep{dong2025stp}, and critic evolution \citep{chen2025spcevolvingselfplaycritic}. SPIRAL \citep{liu2025spiral} proposed multi-turn game training that transfers to mathematical reasoning, while concurrent work explored game-based RL for vision-language models \citep{Xie2025PlayTG,Tong2025GameRLSM,liao2025think}. Foundation models for game agents \citep{magne2025nitrogen,wang2025game} further demonstrate the potential of game environments for capable AI.
\section{Conclusion}

We presented \method{}, a game-based self-play framework that learns transferable reasoning by selectively reinforcing trajectories exhibiting abstract and adaptive reasoning patterns. Starting from the observation that terminal win/loss signals cannot distinguish transferable reasoning from game-specific heuristics, we identified two barriers to transfer: \textit{domain specificity}, addressed by a Reasoning Transferability Coefficient ($\varphi$), and \textit{contextual stasis}, addressed by a Reasoning Evolution Reward ($\psi$). Experiments across mathematical reasoning, general reasoning, and code generation show consistent improvements over base models and SPIRAL, with pronounced gains on competition-level mathematics, and carry over to instruction-tuned backbones. Ablation, human evaluation, and cross-evaluator agreement confirm that \method{} cultivates genuinely abstract and progressive reasoning. More broadly, our results suggest that the structure of the trajectory, not just its outcome, carries the signal that transfers, motivating future work on richer environments, curriculum strategies that compose skills across games, and lightweight local reward models.
\section*{Limitations}
\label{sec:limitations}

Following SPIRAL, we train \method{} on three text-based games from TextArena. While these games provide complementary coverage of core reasoning dimensions, exploring a broader set of game environments, including more complex multi-agent scenarios or games with richer state spaces, may enhance the diversity of learned reasoning patterns. Our experiments cover Qwen3-4B in both base and instruction-tuned variants (\S\ref{sec:instruct_generalization}); scaling to larger backbones and additional model families remains an important direction for understanding how reasoning transfer interacts with model capacity. Finally, $\varphi$ and $\psi$ are currently computed with GPT-4, which introduces an external API dependency; distilling the evaluator into a lightweight local reward model is a natural next step toward fully self-contained training.

\section*{Acknowledgments}

Xiaocheng Feng and Lingpeng Kong are the co-corresponding authors of this work.
We thank the anonymous reviewers for their insightful comments.
This work was supported by the National Natural Science Foundation of China (NSFC) (grant 62522603, 62276078), the Key R\&D Program of Heilongjiang via grant 2022ZX01A32, the Fundamental Research Funds for the Central Universities (XNJKKGYDJ2024013) .

% Custom bibliography entries only
\bibliography{custom}

\appendix

\section{Detailed Experimental Results}
\label{app:detailed_results}

Table~\ref{tab:main_results_full} presents the complete numerical results across all evaluation benchmarks. We compare \method{} against the Qwen3-4B-Base model and SPIRAL, reporting accuracy percentages for each benchmark along with improvement deltas.

\paragraph{Key Observations.} \method{} achieves the highest performance on 8 out of 9 benchmarks. The most substantial gains appear on mathematical reasoning tasks, particularly on competition-level problems (AIME24, AIME25, AMC-23) where strategic thinking and multi-step reasoning are essential. AIME24 shows a 2$\times$ improvement (10.00\% $\rightarrow$ 20.00\%), while AMC-23 improves by 10 percentage points. On Minerva Math, \method{} (41.50\%) slightly trails SPIRAL (42.30\%) but still achieves a 17.2 percentage point improvement over the baseline. On general reasoning benchmarks, \method{} consistently outperforms both the baseline and SPIRAL. HumanEval (pass@1) shows a 10 percentage point improvement over the baseline, demonstrating that game-based training enhances programming capabilities through improved logical structuring.

\begin{table*}[h]
\centering
\resizebox{\textwidth}{!}{
\begin{tabular}{l|cccccc|cc|c}
\toprule
\multirow{2}{*}{\textbf{Model}} & \multicolumn{6}{c|}{\textbf{Mathematical Reasoning}} & \multicolumn{2}{c|}{\textbf{General}} & \textbf{Code} \\
& MATH500 & AIME24 & AIME25 & OlympiadBench & AMC-23 & Minerva & GPQA & MMLU-Pro & HumanEval \\
\midrule
Qwen3-4B-Base & 65.80 & 10.00 & 3.30 & 33.30 & 50.00 & 24.30 & 30.60 & 47.20 & 67.93 \\
SPIRAL & 71.00 & 10.00 & 6.70 & 34.70 & 45.00 & 42.30 & 36.41 & 53.93 & 77.44 \\
\textbf{\method{} (Ours)} & \textbf{76.00} & \textbf{20.00} & \textbf{13.30} & \textbf{39.90} & \textbf{60.00} & 41.50 & \textbf{38.23} & \textbf{57.83} & \textbf{77.93} \\
\midrule
$\Delta$ vs. Base & \textcolor{blue}{+10.20} & \textcolor{blue}{+10.00} & \textcolor{blue}{+10.00} & \textcolor{blue}{+6.60} & \textcolor{blue}{+10.00} & \textcolor{blue}{+17.20} & \textcolor{blue}{+7.63} & \textcolor{blue}{+10.63} & \textcolor{blue}{+10.00} \\
$\Delta$ vs. SPIRAL & \textcolor{blue}{+5.00} & \textcolor{blue}{+10.00} & \textcolor{blue}{+6.60} & \textcolor{blue}{+5.20} & \textcolor{blue}{+15.00} & \textcolor{red}{-0.80} & \textcolor{blue}{+1.82} & \textcolor{blue}{+3.90} & \textcolor{blue}{+0.49} \\
\bottomrule
\end{tabular}}
\caption{Complete benchmark results. All values are accuracy percentages. Best results in each column are \textbf{bolded}. $\Delta$ rows show improvement over baseline and SPIRAL respectively. \textcolor{blue}{Blue} indicates improvement; \textcolor{red}{red} indicates regression.}
\label{tab:main_results_full}
\end{table*}

\section{Ablation Study Details}
\label{app:ablation}

Table~\ref{tab:ablation_full} presents the complete ablation study comparing the full \method{} framework against its variant without the Reasoning Evolution Reward ($\psi$). This ablation isolates the contribution of $\psi$, which captures the dynamic quality of reasoning development across game trajectories.

\paragraph{Detailed Analysis.} The results reveal that $\psi$ provides consistent benefits across nearly all benchmarks. Removing $\psi$ causes substantial degradation on competition-level mathematical reasoning: AIME24 drops by 6.70\% (from 20.00\% to 13.30\%) and AMC-23 by 7.50\% (from 60.00\% to 52.50\%). AIME25 decreases by 3.30\%, and MATH500 by 1.40\%. General reasoning tasks also benefit: GPQA improves by 1.01\% and MMLU-Pro by 0.91\% with $\psi$. The only exception is Minerva Math, where $\psi$ leads to a slight decrease of 1.10\%. This pattern confirms that $\psi$ is particularly valuable for tasks requiring extended multi-step reasoning and strategic adaptation, precisely the capabilities that the Reasoning Evolution Reward is designed to incentivize. The consistent improvements across 8 out of 9 benchmarks demonstrate that capturing reasoning evolution is essential for robust transfer learning.

\begin{table*}[h]
\centering
\resizebox{\textwidth}{!}{
\begin{tabular}{l|cccccc|cc|c}
\toprule
\multirow{2}{*}{\textbf{Model}} & \multicolumn{6}{c|}{\textbf{Mathematical Reasoning}} & \multicolumn{2}{c|}{\textbf{General}} & \textbf{Code} \\
& MATH500 & AIME24 & AIME25 & OlympiadBench & AMC-23 & Minerva & GPQA & MMLU-Pro & HumanEval \\
\midrule
\textbf{\method{} (Full)} & \textbf{76.00} & \textbf{20.00} & \textbf{13.30} & \textbf{39.90} & \textbf{60.00} & 41.50 & \textbf{38.23} & \textbf{57.83} & \textbf{77.93} \\
\method{} (w/o $\psi$) & 74.60 & 13.30 & 10.00 & 39.30 & 52.50 & \textbf{42.60} & 37.22 & 56.92 & 77.80 \\
\midrule
$\Delta$ (Full $-$ w/o $\psi$) & \textcolor{blue}{+1.40} & \textcolor{blue}{+6.70} & \textcolor{blue}{+3.30} & \textcolor{blue}{+0.60} & \textcolor{blue}{+7.50} & \textcolor{red}{-1.10} & \textcolor{blue}{+1.01} & \textcolor{blue}{+0.91} & \textcolor{blue}{+0.13} \\
\bottomrule
\end{tabular}}
\caption{Ablation study: Impact of Reasoning Evolution Reward ($\psi$). Best results per column are \textbf{bolded}. $\Delta$ row shows the contribution of $\psi$ (positive values indicate $\psi$ improves performance). \textcolor{blue}{Blue} indicates $\psi$ helps; \textcolor{red}{red} indicates $\psi$ hurts.}
\label{tab:ablation_full}
\end{table*}

\section{Parameter Sensitivity Analysis}
\label{app:beta_sensitivity}

Table~\ref{tab:beta_sensitivity} presents the complete parameter sensitivity analysis for the Reasoning Evolution Reward coefficient $\beta$. We evaluate five values spanning two orders of magnitude ($\beta \in \{0.01, 0.05, 0.10, 0.20, 0.30\}$) to understand how this hyperparameter affects downstream reasoning transfer.

\paragraph{Key Findings.} The results reveal a clear optimal region around $\beta = 0.20$, which achieves the best performance on 6 out of 9 benchmarks. The sensitivity analysis yields several insights:

\begin{itemize}[leftmargin=*, nosep]
    \item \textbf{Robustness in the moderate range:} Performance remains relatively stable for $\beta \in [0.10, 0.20]$, suggesting that the method is not highly sensitive to precise hyperparameter tuning within this range.

    \item \textbf{Under-weighting effects:} At $\beta = 0.01$, the reasoning evolution signal has minimal impact, and results approximate those of the ablated model without $\psi$. This confirms that the $\beta$ coefficient effectively controls the contribution of the reasoning evolution reward.

    \item \textbf{Over-weighting effects:} At $\beta = 0.30$, several benchmarks show substantial degradation (MATH500: $-4.4\%$, AMC-23: $-12.5\%$, AIME25: $-6.6\%$), indicating that excessive emphasis on reasoning evolution metrics can interfere with the primary game-based learning objective.

    \item \textbf{Task-specific preferences:} Competition-level mathematics (AIME24) shows continued improvement up to $\beta = 0.30$, while science-focused tasks (Minerva Math) peak at lower values ($\beta = 0.10$). This suggests that different reasoning domains may benefit from different $\beta$ settings, though $\beta = 0.20$ provides the best overall balance.
\end{itemize}

\begin{table*}[h]
\centering
\resizebox{\textwidth}{!}{
\begin{tabular}{l|cccccc|cc|c}
\toprule
\multirow{2}{*}{\textbf{$\beta$}} & \multicolumn{6}{c|}{\textbf{Mathematical Reasoning}} & \multicolumn{2}{c|}{\textbf{General}} & \textbf{Code} \\
& MATH500 & AIME24 & AIME25 & OlympiadBench & AMC-23 & Minerva & GPQA & MMLU-Pro & HumanEval \\
\midrule
0.01 & 75.80 & 16.70 & 10.00 & 37.20 & 57.50 & 37.90 & 37.22 & 53.13 & 76.59 \\
0.05 & 75.60 & 10.00 & 13.30 & 37.80 & 57.50 & 39.70 & 36.01 & 54.60 & 77.20 \\
0.10 & 75.60 & 13.30 & 13.30 & 37.00 & 57.50 & \textbf{46.00} & 36.16 & 55.09 & 77.68 \\
\rowcolor{green!10} \textbf{0.20} & \textbf{76.00} & 20.00 & \textbf{13.30} & \textbf{39.90} & \textbf{60.00} & 41.50 & \textbf{38.23} & \textbf{57.83} & \textbf{77.93} \\
0.30 & 71.60 & \textbf{23.30} & 6.70 & 36.10 & 47.50 & 34.90 & 34.80 & 56.51 & 77.44 \\
\bottomrule
\end{tabular}}
\caption{Parameter sensitivity analysis for the Reasoning Evolution Reward coefficient $\beta$. All values are accuracy percentages. Best results per column are \textbf{bolded}. The optimal setting $\beta = 0.20$ (highlighted in green) achieves the best overall performance across benchmark categories.}
\label{tab:beta_sensitivity}
\end{table*}

\section{Prompt Templates}
\label{app:prompts}

This section presents all prompt templates used in training and evaluation. We organize them into three categories: training prompts for self-play (\S\ref{app:training_prompts}), evaluation prompts for computing $\varphi$ and $\psi$ (\S\ref{app:tam_prompts}), and benchmark evaluation prompts (\S\ref{app:eval_prompts}).

\subsection{Training Prompts}
\label{app:training_prompts}

We use two prompt templates during training: one for game self-play (Figure~\ref{fig:selfplay_prompt}) and one for online mathematical reasoning evaluation (Figure~\ref{fig:general_prompt}).

\begin{figure}[h]
\begin{tcolorbox}[
    colback=gray!5,
    colframe=gray!60!black,
    title={\textbf{Self-Play Game Prompt}},
    fonttitle=\bfseries\footnotesize,
    boxrule=0.5pt,
    arc=2mm,
    left=4pt,
    right=4pt,
    top=2pt,
    bottom=2pt,
    width=\columnwidth
]
{\small
\texttt{<|im\_start|>user}\\
You are playing a two-player zero-sum game. Make valid actions to win.\\
Observation: \texttt{\{observation\}}\\
Please reason step by step, and put your final answer within \texttt{\textbackslash boxed\{\}}.\texttt{<|im\_end|>}\\
\texttt{<|im\_start|>assistant}
}
\end{tcolorbox}
\caption{Prompt template for game self-play training and online game evaluation. The \texttt{\{observation\}} placeholder is replaced with the current game state.}
\label{fig:selfplay_prompt}
\end{figure}

\begin{figure}[h]
\begin{tcolorbox}[
    colback=gray!5,
    colframe=gray!60!black,
    title={\textbf{General Reasoning Prompt}},
    fonttitle=\bfseries\footnotesize,
    boxrule=0.5pt,
    arc=2mm,
    left=4pt,
    right=4pt,
    top=2pt,
    bottom=2pt,
    width=\columnwidth
]
{\small
\texttt{<|im\_start|>user}\\
Question: \texttt{\{question\}}\\
Please reason step by step, and put your final answer within \texttt{\textbackslash boxed\{\}}.\texttt{<|im\_end|>}\\
\texttt{<|im\_start|>assistant}
}
\end{tcolorbox}
\caption{Prompt template for online mathematical reasoning evaluation during training (e.g., AIME problems).}
\label{fig:general_prompt}
\end{figure}

\subsection{Trajectory Modulation Prompts}
\label{app:tam_prompts}

The Reasoning Transferability Coefficient ($\varphi$) and Reasoning Evolution Reward ($\psi$) are computed using GPT-4 as the evaluation backbone. We present the complete prompts with detailed scoring criteria.

\subsubsection{Reasoning Transferability Coefficient Prompt}
\label{app:rtc_prompt}

The Reasoning Transferability Coefficient measures whether reasoning patterns in a game trajectory can generalize to other domains such as mathematics and coding. Figure~\ref{fig:rtc_prompt} presents the complete prompt template, which evaluates three dimensions, each scored from 0 to 1.

\begin{figure}[h]
\begin{tcolorbox}[
    colback=orange!5,
    colframe=orange!60!black,
    title={\textbf{RTC Evaluation Prompt}},
    fonttitle=\bfseries\footnotesize,
    boxrule=0.5pt,
    arc=2mm,
    left=4pt,
    right=4pt,
    top=2pt,
    bottom=2pt,
    width=\columnwidth
]
{\scriptsize
You are a professional reasoning transferability expert. Your task is to evaluate whether game reasoning is transferable (i.e., applicable to other domains like mathematics and coding).

\textbf{Background Knowledge}

\textbf{Transferable reasoning}: Uses abstract concepts, structured frameworks, and general principles applicable to other domains.
\begin{itemize}[leftmargin=*,nosep]
\item Example: ``Enumerate all cases $\rightarrow$ compute expected payoff for each $\rightarrow$ select optimal'' (applicable to any decision problem)
\end{itemize}

\textbf{Non-transferable reasoning}: Relies on game-specific terminology, experiential memory, or concrete patterns only valid in the current game.
\begin{itemize}[leftmargin=*,nosep]
\item Example: ``King usually wins, so bet'' (only valid in Poker)
\end{itemize}

\textbf{Game Trajectory}

Game: \texttt{\{game\_name\}}

\texttt{\{trajectory\_text\}}

\textbf{Scoring Criteria}

\textbf{Dimension 1: Abstraction Level.} Does reasoning use abstract concepts or game-specific terms?
\begin{itemize}[leftmargin=*,nosep]
\item \textbf{1.0 (High)}: Domain-agnostic concepts (``expected value,'' ``enumerate possibilities,'' ``probability distribution'')
\item \textbf{0.5 (Medium)}: Mix of abstract and game-specific (``King's probability is 1/2, so expected payoff...'')
\item \textbf{0.0 (Low)}: Entirely game-specific (``King beats Queen,'' ``center position is important'')
\end{itemize}

\textbf{Dimension 2: Structural Clarity.} Does reasoning use clear, reusable frameworks?
\begin{itemize}[leftmargin=*,nosep]
\item \textbf{1.0 (High)}: Clear frameworks (case-by-case analysis, EV calculation, if-then chains)
\item \textbf{0.5 (Medium)}: Some structure but incomplete (``I considered several cases...'')
\item \textbf{0.0 (Low)}: Unstructured, arbitrary statements (``I think this is good,'' ``Based on experience...'')
\end{itemize}

\textbf{Dimension 3: Principle Orientation.} Is reasoning based on general principles or game-specific experience?
\begin{itemize}[leftmargin=*,nosep]
\item \textbf{1.0 (High)}: Explicit principles (``by Bayes' theorem,'' ``to maximize expected utility'')
\item \textbf{0.5 (Medium)}: Implicit principles (``I need to balance risk and reward'')
\item \textbf{0.0 (Low)}: Experience-based (``I've seen this position before,'' ``Opponents usually...'')
\end{itemize}

\textbf{Key Judgment}: If game terms are replaced with variables (e.g., ``King'' $\rightarrow$ ``Option A''), does the reasoning logic remain valid and meaningful? If yes $\rightarrow$ high score; if no $\rightarrow$ low score.

\textbf{Output JSON:}
\texttt{\{"abstraction\_level": <0-1>, "structural\_clarity": <0-1>, "principle\_based": <0-1>, "explanation": "<50-100 words>", "key\_transferable\_patterns": ["<pattern1>", "<pattern2>"]\}}
}
\end{tcolorbox}
\caption{Complete prompt template for computing the Reasoning Transferability Coefficient ($\varphi$). The evaluator assesses three dimensions: abstraction level, structural clarity, and principle orientation.}
\label{fig:rtc_prompt}
\end{figure}

\subsubsection{Reasoning Evolution Reward Prompt}
\label{app:rer_prompt}

The Reasoning Evolution Reward captures the quality of reasoning development across a game trajectory. Figure~\ref{fig:rer_prompt} presents the complete prompt template. Each dimension is scored from $-1$ to $+1$, allowing the metric to penalize degradation.

\begin{figure}[h]
\begin{tcolorbox}[
    colback=blue!5,
    colframe=blue!60!black,
    title={\textbf{RER Evaluation Prompt}},
    fonttitle=\bfseries\footnotesize,
    boxrule=0.5pt,
    arc=2mm,
    left=4pt,
    right=4pt,
    top=2pt,
    bottom=2pt,
    width=\columnwidth
]
{\scriptsize
You are a professional reasoning analysis expert. Your task is to evaluate the evolution quality of reasoning across a game trajectory.

\textbf{Game Trajectory}

Game: \texttt{\{game\_name\}}

\texttt{\{trajectory\_text\}}

\textbf{Scoring Criteria}

\textbf{Dimension 1: Reasoning Deepening.} Does reasoning progress from simple to complex?
\begin{itemize}[leftmargin=*,nosep]
\item \textbf{+1}: Progressive deepening (``control center'' $\rightarrow$ ``analyze opponent threats'' $\rightarrow$ ``build dual attack'')
\item \textbf{0}: Constant complexity level (simple action descriptions each turn)
\item \textbf{-1}: Degradation (detailed analysis initially, later just ``I play here'')
\end{itemize}

\textbf{Dimension 2: Strategy Adaptation.} Does reasoning adapt to opponent behavior or game state?
\begin{itemize}[leftmargin=*,nosep]
\item \textbf{+1}: Clear adaptation (``Opponent took the corner, I need to change my plan...'')
\item \textbf{0}: Fixed strategy (executing predetermined plan regardless of opponent)
\item \textbf{-1}: Erratic or contradictory (``I'll attack'' $\rightarrow$ ``I'll defend'' $\rightarrow$ ``I'll attack'' without reason)
\end{itemize}

\textbf{Dimension 3: Logical Coherence.} Does later reasoning build causally on earlier conclusions?
\begin{itemize}[leftmargin=*,nosep]
\item \textbf{+1}: Causal chain (``Because X, I did Y'' $\rightarrow$ ``Y resulted in Z, so next...'')
\item \textbf{0}: Independent but non-contradictory (each turn has reasoning but no cross-references)
\item \textbf{-1}: Contradictory (earlier: ``must defend,'' later: ``should have attacked'')
\end{itemize}

\textbf{Special Cases}:
\begin{itemize}[leftmargin=*,nosep]
\item Trajectory with 1 to 2 turns: default score 0 (cannot evaluate evolution)
\item Empty reasoning (\texttt{<think></think>}): score -1 (reasoning collapse)
\item Very short reasoning (<20 tokens per turn): tend toward negative scores
\end{itemize}

\textbf{Output JSON:}
\texttt{\{"reasoning\_deepening": <-1 to 1>, "strategy\_adaptation": <-1 to 1>, "logical\_coherence": <-1 to 1>, "explanation": "<50-100 words>"\}}
}
\end{tcolorbox}
\caption{Complete prompt template for computing the Reasoning Evolution Reward ($\psi$). The evaluator assesses three dimensions: reasoning deepening, strategy adaptation, and logical coherence.}
\label{fig:rer_prompt}
\end{figure}

\subsection{Benchmark Evaluation Prompts}
\label{app:eval_prompts}

We use three prompt templates for downstream benchmark evaluation: mathematical reasoning (Figure~\ref{fig:math_prompt}), multiple choice (Figure~\ref{fig:mcq_prompt}), and code generation (Figure~\ref{fig:code_prompt}).

\begin{figure}[h]
\begin{tcolorbox}[
    colback=green!5,
    colframe=green!60!black,
    title={\textbf{Mathematical Reasoning Prompt}},
    fonttitle=\bfseries\footnotesize,
    boxrule=0.5pt,
    arc=2mm,
    left=4pt,
    right=4pt,
    top=2pt,
    bottom=2pt,
    width=\columnwidth
]
{\small
\texttt{<|im\_start|>user}\\
Please reason step by step, and put your final answer within \texttt{\textbackslash boxed\{\}}.\\
Question: \texttt{\{input\}}\texttt{<|im\_end|>}\\
\texttt{<|im\_start|>assistant}
}
\end{tcolorbox}
\caption{Prompt template for mathematical reasoning benchmarks (MATH500, AIME, AMC, OlympiadBench, Minerva Math).}
\label{fig:math_prompt}
\end{figure}

\begin{figure}[h]
\begin{tcolorbox}[
    colback=green!5,
    colframe=green!60!black,
    title={\textbf{Multiple Choice Prompt}},
    fonttitle=\bfseries\footnotesize,
    boxrule=0.5pt,
    arc=2mm,
    left=4pt,
    right=4pt,
    top=2pt,
    bottom=2pt,
    width=\columnwidth
]
{\small
Please reason step by step, and put your final answer within \texttt{\textbackslash boxed\{\}}. Your final answer should be of the following format: \texttt{\textbackslash boxed\{LETTER\}} where LETTER is one of ABCD.

Question: \texttt{\{question\}}

Options:\\
A) \texttt{\{A\}}\\
B) \texttt{\{B\}}\\
C) \texttt{\{C\}}\\
D) \texttt{\{D\}}
}
\end{tcolorbox}
\caption{Prompt template for multiple choice benchmarks (GPQA, MMLU-Pro). For MMLU-Pro, options extend to A through J.}
\label{fig:mcq_prompt}
\end{figure}

\begin{figure}[h]
\begin{tcolorbox}[
    colback=green!5,
    colframe=green!60!black,
    title={\textbf{Code Generation Prompt}},
    fonttitle=\bfseries\footnotesize,
    boxrule=0.5pt,
    arc=2mm,
    left=4pt,
    right=4pt,
    top=2pt,
    bottom=2pt,
    width=\columnwidth
]
{\small
Read the following function signature and docstring, and fully implement the function described. Your response should only contain the code for this function.

\texttt{\{function\_signature\_and\_docstring\}}
}
\end{tcolorbox}
\caption{Prompt template for code generation benchmark (HumanEval).}
\label{fig:code_prompt}
\end{figure}

\section{Human Evaluation Details}
\label{app:human_eval}

This section provides complete details of the human evaluation study described in \S\ref{sec:human_eval}, including evaluation guidelines, expert-level breakdowns, and inter-annotator agreement statistics.

\subsection{Evaluation Protocol}

We randomly sample 50 reasoning traces from game trajectories (Kuhn Poker and Tic-Tac-Toe) generated by each of the four models: Qwen3-4B-Base, SPIRAL, \method{} (w/o $\psi$), and \method{}. Five expert annotators (graduate students with backgrounds in NLP and machine learning) independently evaluate each trace. Annotators are blind to model identity and evaluate traces in randomized order.

\subsection{Evaluation Dimensions}

Each trace is scored on a 1 to 5 Likert scale along two dimensions:

\paragraph{Reasoning Abstraction (1 to 5).} This dimension measures the degree to which reasoning employs domain-agnostic, transferable patterns:
\begin{itemize}[leftmargin=*, nosep]
    \item \textbf{1 (Poor):} Reasoning relies entirely on game-specific heuristics (e.g., ``I should bluff because that's what poker players do'').
    \item \textbf{2 (Below Average):} Reasoning is predominantly game-specific with occasional abstract observations that lack development.
    \item \textbf{3 (Moderate):} Reasoning mixes game-specific and abstract concepts in roughly equal proportion.
    \item \textbf{4 (Good):} Reasoning uses mostly abstract concepts with only minor game-specific terminology.
    \item \textbf{5 (Excellent):} Reasoning uses explicit probability calculations, expected value analysis, and systematic case enumeration that would transfer to mathematics or coding.
\end{itemize}

\paragraph{Reasoning Progression (1 to 5).} This dimension measures the dynamic quality of reasoning development:
\begin{itemize}[leftmargin=*, nosep]
    \item \textbf{1 (Poor):} Reasoning is shallow, repetitive, or degrades over time.
    \item \textbf{2 (Below Average):} Reasoning shows minimal development; largely repetitive with occasional improvements.
    \item \textbf{3 (Moderate):} Reasoning maintains consistency but does not deepen substantially.
    \item \textbf{4 (Good):} Reasoning shows clear development and adaptation with minor inconsistencies.
    \item \textbf{5 (Excellent):} Reasoning progressively deepens, adapts to new information, and builds coherently on earlier conclusions.
\end{itemize}

\subsection{Aggregated Results}

Table~\ref{tab:human_eval_summary} presents the mean scores and standard errors across all annotators and samples.

\begin{table}[h]
\centering
\resizebox{\columnwidth}{!}{
\begin{tabular}{l|cc}
\toprule
\textbf{Model} & \textbf{Abstraction} & \textbf{Progression} \\
\midrule
Qwen3-4B-Base & 2.48 $\pm$ 0.08 & 2.32 $\pm$ 0.09 \\
SPIRAL & 3.24 $\pm$ 0.07 & 3.08 $\pm$ 0.08 \\
\method{} (w/o $\psi$) & 3.82 $\pm$ 0.06 & 3.36 $\pm$ 0.07 \\
\textbf{\method{}} & \textbf{4.06 $\pm$ 0.05} & \textbf{4.18 $\pm$ 0.06} \\
\bottomrule
\end{tabular}}
\caption{Human evaluation scores (mean $\pm$ SE) on 1 to 5 scale.}
\label{tab:human_eval_summary}
\end{table}

\section{Additional Case Studies}
\label{app:case_study}

This section presents two additional case studies complementing the Tic-Tac-Toe analysis in \S\ref{sec:case_study}. These cases further demonstrate how \method{}'s trajectory advantage modulation improves reasoning abstraction ($\varphi$) and progression ($\psi$) across different game types.

\subsection{Case Study: Kuhn Poker}
\label{app:case_kuhn}

Kuhn Poker requires probabilistic reasoning and strategic deception, making it an ideal testbed for evaluating abstract reasoning capabilities. Table~\ref{tab:case_kuhn} compares reasoning traces from the baseline and \method{} during a five-round game.

\begin{table*}[h]
\centering
\small
\resizebox{\textwidth}{!}{
\begin{tabular}{p{0.12\textwidth}|p{0.40\textwidth}|p{0.40\textwidth}}
\toprule
\textbf{Turn} & \textbf{Baseline} & \textbf{\method{}} \\
\midrule
\textbf{Turn 1}
\newline Card: J
\newline Action: [bet]
&
``You have the King (K), which is the highest card... This is a strong hand.''
\newline\newline
\textcolor{baselinecolor}{\ding{56} \textit{Hallucination: Misidentifies J as K}}
&
``Folding early can sometimes be a \textcolor{abstractcolor}{\textbf{strategic move to bluff later}} or to \textcolor{abstractcolor}{\textbf{control the pace of the game}}.''
\newline\newline
\textcolor{abstractcolor}{\ding{52} \textit{Abstract: Long-term strategic thinking}}
\\
\midrule
\textbf{Turn 3}
\newline Card: K
\newline Opp: [bet]
&
``At this point, I have 0 chips after Round 2... If I [call]... I might be in a second-place position.''
\newline\newline
\textcolor{progresscolor}{\ding{52} \textit{Progression: Tracks game history}}
&
``Since the game is \textcolor{abstractcolor}{\textbf{zero-sum}} and both players are \textcolor{abstractcolor}{\textbf{rational}}... bluffing is only effective if there is uncertainty.''
\newline\newline
\textcolor{abstractcolor}{\ding{52} \textit{Abstract: Game-theoretic reasoning}}
\\
\midrule
\textbf{Turn 5}
\newline Card: J
\newline Opp: [bet]
&
``My Q is higher than the opponent's last known hand... reasonable to [bet].''
\newline\newline
\textcolor{baselinecolor}{\ding{56} \textit{Hallucination: Confuses J with Q}}
&
``\textcolor{abstractcolor}{\textbf{Expected Value Analysis}}: Folding: 0 expected loss. Calling: risk 2 chips...''
\newline\newline
\textcolor{abstractcolor}{\ding{52} \textit{Abstract: Formal EV calculation}}
\\
\bottomrule
\end{tabular}}
\caption{Kuhn Poker case study comparing baseline and \method{} reasoning. \method{} demonstrates abstract game-theoretic concepts (\textcolor{abstractcolor}{blue}), while the baseline shows stronger state tracking (\textcolor{progresscolor}{green}) but suffers from card hallucinations (\textcolor{baselinecolor}{red}).}
\label{tab:case_kuhn}
\end{table*}

\paragraph{Analysis.} The Kuhn Poker case reveals an interesting pattern: \method{} excels in \textit{abstraction} by employing game-theoretic terminology (``zero-sum,'' ``rational,'' ``expected value'') that directly transfers to mathematical reasoning. The baseline, while occasionally tracking game history correctly (Turn 3), suffers from critical perception errors (hallucinating incorrect cards in Turn 1 and 5), which undermines its reasoning coherence. \method{}'s use of formal frameworks (``enumerate all cases $\rightarrow$ compute expected payoff'') mirrors the systematic analysis required for competition-level mathematics.

\subsection{Case Study: Negotiation}
\label{app:case_negotiation}

The Negotiation game requires theory of mind reasoning, value assessment, and strategic communication. Table~\ref{tab:case_negotiation} contrasts reasoning patterns across a multi-turn negotiation.

\begin{table*}[h]
\centering
\small
\resizebox{\textwidth}{!}{
\begin{tabular}{p{0.12\textwidth}|p{0.40\textwidth}|p{0.40\textwidth}}
\toprule
\textbf{Turn} & \textbf{Baseline} & \textbf{\method{}} \\
\midrule
\textbf{Turn 1}
\newline Resources:
\newline Wood: 10 (v=5)
\newline Gold: 10 (v=15)
&
``Wood: 10, Value: 5; Gold: 10, Value: 15. Total: \textcolor{baselinecolor}{\textbf{5 + 15 = 20}}.''
\newline\newline
\textcolor{baselinecolor}{\ding{56} \textit{Low abstraction: Basic arithmetic only}}
&
``Player 0 wants to \textcolor{abstractcolor}{\textbf{strengthen his position}} and \textcolor{abstractcolor}{\textbf{demonstrate willingness to negotiate}}.''
\newline\newline
\textcolor{abstractcolor}{\ding{52} \textit{Abstract: Strategic intent modeling}}
\\
\midrule
\textbf{Turn 3}
\newline P0 offers:
\newline 3 Wood → 10 Gold
&
``A logical trade offer... giving up lower value for higher value.''
\newline\newline
\textcolor{baselinecolor}{\ding{56} \textit{Stateless: No reference to prior turns}}
&
``P1's Initial Offer: 2 Wood → 5 Gold vs. \textcolor{progresscolor}{\textbf{P0's Current Requirement}}: 3 Wood → 10 Gold.''
\newline\newline
\textcolor{progresscolor}{\ding{52} \textit{Progression: Compares offer history}}
\\
\midrule
\textbf{Turn 5}
\newline Negotiation
\newline continues
&
``Given your strong position... the best answer is \$\textbackslash boxed\{1\}\$.''
\newline\newline
\textcolor{baselinecolor}{\ding{56} \textit{Incoherent: Output doesn't follow format}}
&
``I understand your desire, but I can only offer 2 Wood for 5 Gold. This will increase my total value.''
\newline\newline
\textcolor{progresscolor}{\ding{52} \textit{Progression: Reiterates position strategically}}
\\
\bottomrule
\end{tabular}}
\caption{Negotiation case study comparing baseline and \method{} reasoning. \method{} demonstrates psychological modeling (\textcolor{abstractcolor}{blue}) and strategic position maintenance (\textcolor{progresscolor}{green}), while the baseline shows arithmetic-only thinking (\textcolor{baselinecolor}{red}).}
\label{tab:case_negotiation}
\end{table*}

\paragraph{Analysis.} The Negotiation case most clearly demonstrates the difference between arithmetic-level and strategic-level reasoning. The baseline treats negotiation as a simple value calculation problem, computing ``5 + 15 = 20'' and making greedy offers. \method{}, by contrast, models opponent intent (``wants to strengthen position''), tracks negotiation history (``Initial Offer vs.\ Current Requirement''), and strategically maintains positions through reiteration. These capabilities (theory of mind, historical context, and strategic communication) are precisely the skills that transfer to complex mathematical word problems requiring multiple constraint satisfaction.

\paragraph{Summary.} Across all three game types, \method{} addresses the two fundamental challenges: abstract domain-agnostic concepts overcome \textit{domain specificity} ($\varphi$), while progressive state-aware reasoning overcomes \textit{contextual stasis} ($\psi$). The baseline exhibits characteristic failure modes reflecting these challenges: game-specific heuristics (domain specificity), reset issues treating each turn as independent (contextual stasis), and arithmetic-only thinking lacking strategic abstraction. These patterns explain why \method{}'s targeted approach produces superior transfer to mathematical reasoning benchmarks.

\section{Task Formulation Background}
\label{app:task_formulation}

This section provides extended background on the formal frameworks underlying our approach: Markov Decision Processes, their turn-level extensions, and two-player zero-sum Markov games.

\subsection{Markov Decision Processes}
\label{app:mdp}

A Markov Decision Process (MDP) provides the foundational framework for sequential decision-making under uncertainty. Formally, an MDP is defined as a tuple $\mathcal{M} = (\mathcal{S}, \mathcal{A}, T, r, \gamma)$ where:

\begin{itemize}[leftmargin=*, nosep]
    \item $\mathcal{S}$: The state space, representing all possible configurations of the environment
    \item $\mathcal{A}$: The action space, representing all possible decisions the agent can make
    \item $T: \mathcal{S} \times \mathcal{A} \times \mathcal{S} \rightarrow [0, 1]$: The transition function, where $T(s' | s, a)$ gives the probability of transitioning to state $s'$ when taking action $a$ in state $s$
    \item $r: \mathcal{S} \times \mathcal{A} \rightarrow \mathbb{R}$: The reward function, mapping state-action pairs to scalar rewards
    \item $\gamma \in [0, 1]$: The discount factor, balancing immediate versus future rewards
\end{itemize}

The agent's goal is to learn a policy $\pi: \mathcal{S} \rightarrow \Delta(\mathcal{A})$ that maximizes the expected cumulative discounted reward:
\begin{equation}
J(\pi) = \mathbb{E}_{\tau \sim \pi}\left[\sum_{t=0}^{\infty} \gamma^t r(s_t, a_t)\right]
\end{equation}
where $\tau = (s_0, a_0, s_1, a_1, \ldots)$ denotes a trajectory sampled by following policy $\pi$.

\subsection{Turn-Level MDPs for Language Models}
\label{app:turn_level_mdp}

Standard MDPs operate at the token level for language models, where each action corresponds to generating a single token. However, this formulation presents challenges for multi-turn reasoning:

\begin{enumerate}[leftmargin=*, nosep]
    \item \textbf{Credit assignment}: Rewards are typically sparse (given only at episode end), making it difficult to attribute credit across thousands of tokens
    \item \textbf{Temporal abstraction}: Meaningful reasoning units span multiple tokens, but token-level optimization lacks this structure
    \item \textbf{Computational cost}: Optimizing at the token level requires gradient computation through entire sequences
\end{enumerate}

We address these challenges by formulating a \textit{turn-level MDP}, where actions correspond to complete responses rather than individual tokens. In this formulation:

\begin{itemize}[leftmargin=*, nosep]
    \item \textbf{States} $s_t \in \mathcal{S}$ represent complete interaction contexts, including the problem specification, conversation history, and current game configuration
    \item \textbf{Actions} $a_t \in \mathcal{A}$ are full model responses, each containing reasoning trace $c_t$ and executable action component $a_t^{\text{exec}}$
    \item \textbf{Transitions} $T(s_{t+1} | s_t, a_t)$ are determined by appending the response to the context and updating the environment state
\end{itemize}

The turn-level formulation preserves semantic coherence: each ``action'' represents a complete thought, enabling more meaningful optimization signals. The policy $\pi_\theta(y_t | s_t)$ generates the full response $y_t$ autoregressively but is optimized at the turn level.

\subsection{Two-Player Zero-Sum Markov Games}
\label{app:markov_game}

For competitive multi-agent scenarios, we extend MDPs to Markov games \citep{littman1994markov}. A two-player zero-sum Markov game is defined as $\mathcal{G} = (\mathcal{S}, \mathcal{A}_0, \mathcal{A}_1, T, r, \gamma)$ where:

\begin{itemize}[leftmargin=*, nosep]
    \item $\mathcal{S}$: Shared state space observable by both players
    \item $\mathcal{A}_p$: Action space for player $p \in \{0, 1\}$
    \item $T: \mathcal{S} \times \mathcal{A}_0 \times \mathcal{A}_1 \times \mathcal{S} \rightarrow [0, 1]$: Transition function depending on both players' actions
    \item $r: \mathcal{S} \times \mathcal{A}_0 \times \mathcal{A}_1 \rightarrow \mathbb{R}$: Reward for Player 0 (Player 1 receives $-r$)
    \item $\gamma$: Discount factor
\end{itemize}

The zero-sum property ensures that one player's gain is exactly the other's loss:
\begin{equation}
    \small
r_0(s, a^{(0)}, a^{(1)}) + r_1(s, a^{(0)}, a^{(1)}) = 0 \quad \forall s, a^{(0)}, a^{(1)}
\end{equation}

This creates a natural curriculum: as the policy improves, so does its opponent (since both players share the same policy), continuously providing challenging training signal. The Nash equilibrium concept extends naturally: a pair of policies $(\pi_0^*, \pi_1^*)$ is a Nash equilibrium if neither player can improve by unilaterally deviating.

\paragraph{Alternating Turn Structure.} In our formulation, players take turns rather than acting simultaneously. At turn $t$, only player $p = t \bmod 2$ acts, while the other player's action is null. This simplifies the transition dynamics:
\begin{equation}
s_{t+1} = T(s_t, a_t^{(p)}) \quad \text{where } p = t \bmod 2
\end{equation}

The alternating structure naturally models games like chess, Go, and the strategic games in our training suite (Tic-Tac-Toe, Kuhn Poker).

\section{Training Settings Details}
\label{app:training_settings}

This section provides complete hyperparameter configurations for reproducing our experiments.

\paragraph{Optimization Configuration.} Training proceeds for 400 steps with 128 samples per step, yielding 51,200 game transitions total. We use Adam~\citep{kingma2014adam} with learning rate $1 \times 10^{-6}$, batch size 128, and discount factor $\gamma = 1.0$. For role-conditioned advantage estimation, we set EMA decay $\alpha = 0.95$. Trajectories are sampled at temperature $\tau = 1.0$ to encourage exploration.

\paragraph{\method{}-Specific Parameters.} We set the Reasoning Evolution Reward coefficient $\beta = 0.2$ (Equation~\ref{eq:mod_advantage}). The Reasoning Transferability Coefficient $\varphi$ and Reasoning Evolution Reward $\psi$ are computed using GPT-4 as the evaluation backbone, with prompts detailed in \S\ref{app:tam_prompts}.

\paragraph{Computational Resources.} All experiments run on 2 NVIDIA A100 GPUs (80GB) using a distributed actor-learner architecture. Actors generate self-play trajectories using vLLM~\citep{kwon2023efficient} for efficient inference. Each full training run completes in approximately 30 hours.

\section{Game Environment Details}
\label{app:game_envs}

This section provides detailed descriptions of the three text-based zero-sum games used for training.

\paragraph{Tic-Tac-Toe.} A classic $3 \times 3$ grid game serving as our testbed for \textit{spatial reasoning}. Players alternate placing marks to form horizontal, vertical, or diagonal lines of three. The game requires pattern recognition, anticipating opponent moves, and multi-step forcing sequences. As a deterministic perfect-information game, Tic-Tac-Toe isolates pure strategic reasoning from uncertainty management.

\paragraph{Kuhn Poker.} A simplified poker variant~\citep{kuhn1950simplified} emphasizing \textit{probabilistic reasoning}. The game uses only three cards (Jack, Queen, King), where each player receives one card and must decide whether to bet, call, or fold based on incomplete information. Success demands probability estimation, opponent modeling, and expected value calculation under uncertainty.

\paragraph{Simple Negotiation.} A resource trading game developing \textit{strategic optimization} skills. Two players exchange Wood and Gold tokens under opposing utility functions, creating natural tension between competing objectives. Players must infer opponent preferences, plan multi-step trade sequences, and communicate strategically through proposals.

\section{Out-of-Distribution Evaluation Games}
\label{app:ood_games}

We evaluate generalization to games never seen during training. Each OOD game is designed to test whether specific cognitive skills from training games transfer to novel mechanics.

\paragraph{Snake.} A dynamic spatial reasoning game where two players control snakes on a grid, competing to collect apples while avoiding collisions with walls, themselves, or opponents. This tests whether static pattern recognition from Tic-Tac-Toe transfers to trajectory planning and dynamic obstacle avoidance in a real-time environment.

\paragraph{Pig Dice.} A risk-reward decision making game where players repeatedly roll dice to accumulate points but lose all turn points when rolling 1. Players must decide when to ``bank'' accumulated points versus continuing to roll. This tests whether probabilistic reasoning from Kuhn Poker extends to sequential risk assessment and expected value calculation in different contexts.

\paragraph{Truth and Deception.} An asymmetric information game where one player (the Deceiver) knows the true fact among several options and attempts to mislead through conversation, while the other player (the Guesser) must identify truth through strategic questioning. This evaluates whether negotiation skills transfer to pure communication strategy under information asymmetry.

\section{SPIRAL Framework Details}
\label{app:spiral}

This section provides an extended introduction to SPIRAL \citep{liu2025spiral}, the self-play reinforcement learning framework that serves as the foundation for our method.

\subsection{Overview}

SPIRAL (Self-Play on Zero-Sum Games Incentivizes Reasoning via Multi-Agent Multi-Turn Reinforcement Learning) trains language models through competitive self-play on strategic games. The key insight is that zero-sum games provide naturally verifiable rewards without requiring external annotators or reward models: a player either wins, loses, or draws, providing unambiguous training signal.

\subsection{Self-Play Training Loop}

SPIRAL's training proceeds as follows:

\begin{enumerate}[leftmargin=*, nosep]
    \item \textbf{Game Sampling}: Sample a game $G \sim \mathcal{G}$ from the game distribution
    \item \textbf{Trajectory Generation}: Two instances of the current policy $\pi_\theta$ play against each other, generating trajectory $\tau = \{(s_t, y_t^{(p)})\}_{t=0}^T$
    \item \textbf{Outcome Determination}: The game engine determines the winner, assigning rewards $R_p(\tau) \in \{-1, 0, +1\}$
    \item \textbf{Policy Update}: Update $\theta$ using policy gradient with role-conditioned advantages
\end{enumerate}

The self-play mechanism ensures automatic curriculum learning: as the policy improves, its opponent (itself) also improves, maintaining a challenging training distribution throughout learning.

\subsection{Role-Conditioned Advantage Estimation}

A critical challenge in two-player games is that the expected return differs by role. For example, in Tic-Tac-Toe, Player 0 (moving first) has structural advantage. Naively using the same baseline for both players leads to biased gradients.

SPIRAL addresses this through Role-conditioned Advantage Estimation (RAE), maintaining separate baselines $b_{G,p}$ for each game-role pair $(G, p)$:
\begin{equation}
A_{G,p}(\tau) = R_p(\tau) - b_{G,p}
\end{equation}

The baseline is updated via exponential moving average:
\begin{equation}
b_{G,p} \leftarrow \alpha \cdot b_{G,p} + (1 - \alpha) \cdot R_p(\tau)
\end{equation}
where $\alpha$ is the smoothing coefficient (typically 0.99).

\subsection{Policy Gradient Formulation}

The policy gradient for SPIRAL aggregates over all turns played by each role:
\begin{equation}
    \scriptsize
\nabla_\theta J = \mathbb{E}_{G, \tau}\left[\sum_{p \in \{0, 1\}} \sum_{t \in \mathcal{T}_p} A_{G,p}(\tau) \nabla_\theta \log \pi_\theta(y_t^{(p)} | s_t, p, G)\right]
\end{equation}
where $\mathcal{T}_p = \{t : t \bmod 2 = p\}$ indexes the turns belonging to player $p$.

The role conditioning is implemented by prepending a role identifier to the prompt, enabling a single policy to model both players' behavior while accounting for role-specific strategic considerations.

\subsection{Limitations and Motivation for \method{}}

While SPIRAL demonstrates that game-based self-play can improve reasoning, transferring these capabilities to domains like mathematics and coding faces two fundamental challenges:

\begin{enumerate}[leftmargin=*, nosep]
    \item \textbf{Domain Specificity}: SPIRAL optimizes for game outcomes without explicitly encouraging abstract reasoning patterns. Winning strategies often rely on game-specific heuristics (e.g., ``King beats Queen'') rather than domain-agnostic patterns (e.g., ``enumerate cases and compute expected value'').

    \item \textbf{Contextual Stasis}: Games present static problem contexts where rules and settings remain fixed throughout interaction. SPIRAL does not incentivize reasoning that adapts to evolving contexts, yet real-world problems (e.g., mathematical proofs, code debugging) require continuous adaptation as intermediate results reshape the solution space.
\end{enumerate}

These challenges fundamentally limit reasoning transfer. To incentivize transferable reasoning, \method{} addresses both challenges through trajectory advantage modulation: $\varphi$ overcomes domain specificity by measuring abstraction level, while $\psi$ overcomes contextual stasis by rewarding adaptive reasoning development.

\end{document}